\documentclass[lettersize,journal]{IEEEtran}
\usepackage{amsmath,amsfonts}
\usepackage{algorithmic}
\usepackage{algorithm}
\usepackage{array}
\usepackage[caption=false,font=normalsize,labelfont=rm,textfont=rm,labelformat=simple]{subfig}
\usepackage{textcomp}
\usepackage{stfloats}
\usepackage{url}
\usepackage{verbatim}
\usepackage{graphicx}
\usepackage{cite}

\usepackage{xcolor}
\usepackage{lineno}
\usepackage[T1]{fontenc}
\usepackage{enumitem}
\usepackage{amsmath}
\usepackage{multirow}
\usepackage{booktabs}
\usepackage{float}  

\begin{document}

\title{Matching-Free Depth Recovery from Structured Light}

\author{Zhuohang Yu, Kai Wang, Kun Huang, Juyong Zhang
        % <-this % stops a space
\thanks{This work was supported by the National Natural Science Foundation of China under Grant 62441224 and 62272433, the Fundamental Research Funds for the Central Universities WK0010000090. The numerical calculations in this paper have been done on the supercomputing system in the Supercomputing Center of University of Science and Technology of China. \emph{(Corresponding author: Juyong Zhang.)}}% <-this % stops a space
\thanks{Zhuohang Yu and JuyongZhang are with the Department of Mathematics, University of Science and Technology of China, Hefei 230026, China (e-mail: zjyzh@mail.ustc.edu.cn; juyong@ustc.edu.cn).}%
\thanks{Kun Huang is with the Department of Optics and Optical Engineering, School of Physical Sciences, University of Science and Technology of China, Hefei 230026, China (e-mail: huangk17@ustc.edu.cn).}%
\thanks{Kai Wang is with the China Unicom Digital Technology, Beijing 100176, China (e-mail: wangk115@chinaunicom.cn).}}
% The paper headers

\maketitle

\begin{abstract}
We introduce a novel approach for depth estimation using images obtained from monocular structured light systems. In contrast to many existing methods that depend on image matching, our technique employs a density voxel grid to represent scene geometry. This grid is trained through self-supervised differentiable volume rendering. Our method leverages color fields derived from the projected patterns in structured light systems during the rendering process, facilitating the isolated optimization of the geometry field. This innovative approach leads to faster convergence and high-quality results. Additionally, we integrate normalized device coordinates (NDC), a distortion loss, and a distinctive surface-based color loss to enhance geometric fidelity. Experimental results demonstrate that our method outperforms current matching-based techniques in terms of geometric performance in few-shot scenarios, achieving an approximately 30\% reduction in average estimated depth errors for both synthetic scenes and real-world captured scenes. Moreover, our approach allows for rapid training, being approximately three times faster than previous matching-free methods that utilize implicit representations.
\end{abstract}

\begin{IEEEkeywords}
Structured Light, Depth Reconstruction, Volume Rendering, Voxel Grid.
\end{IEEEkeywords}

\section{Introduction}
\label{sec:intro}

\begin{figure}[t]
	\centering
	%\fbox{\rule{0pt}{2in} \rule{0.9\linewidth}{0pt}}
        \subfloat[]{\includegraphics[width=0.9\linewidth]{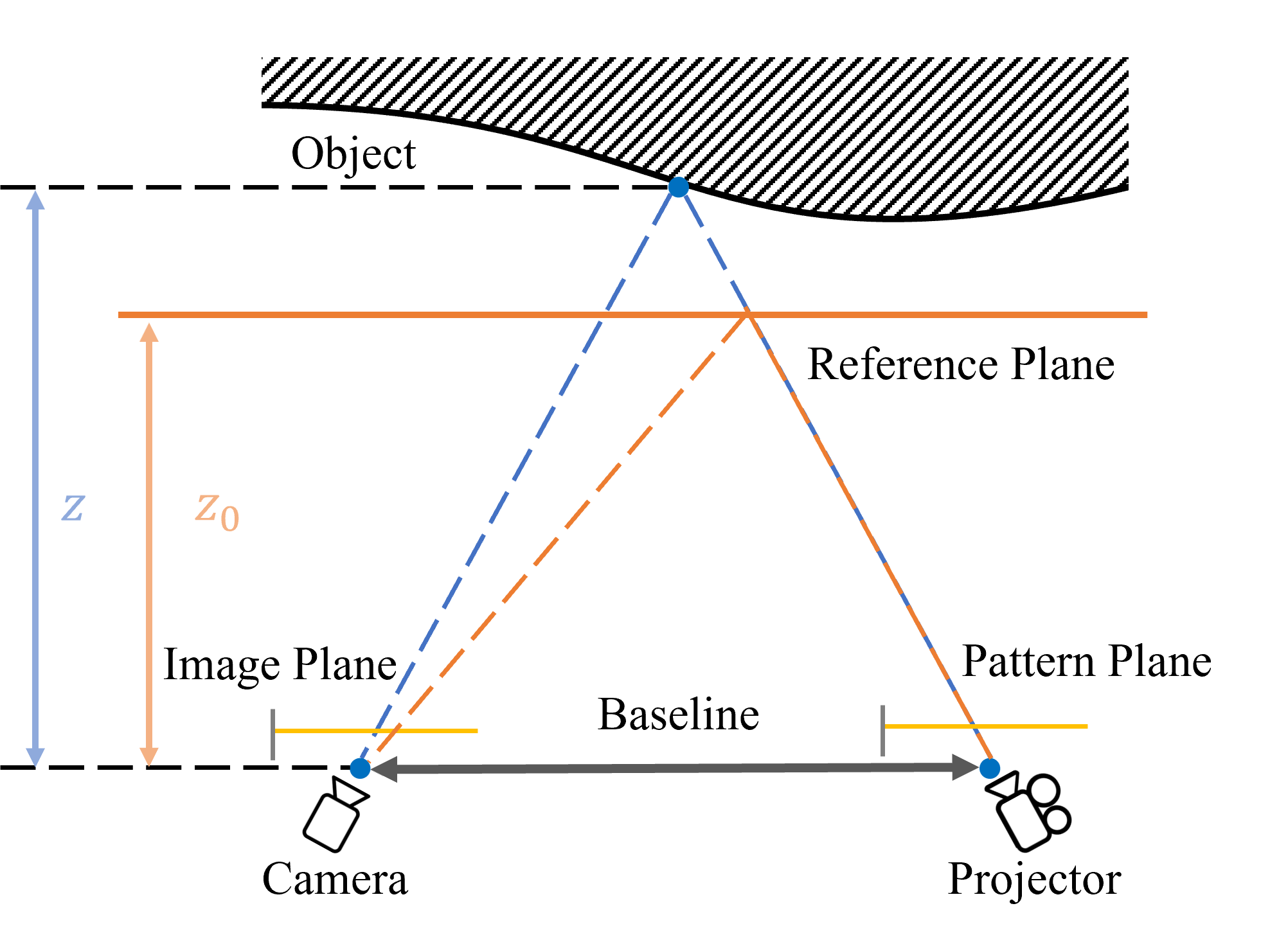}
        \label{fig:monoSL}}
        \\
        \subfloat[]{\includegraphics[width=0.9\linewidth]{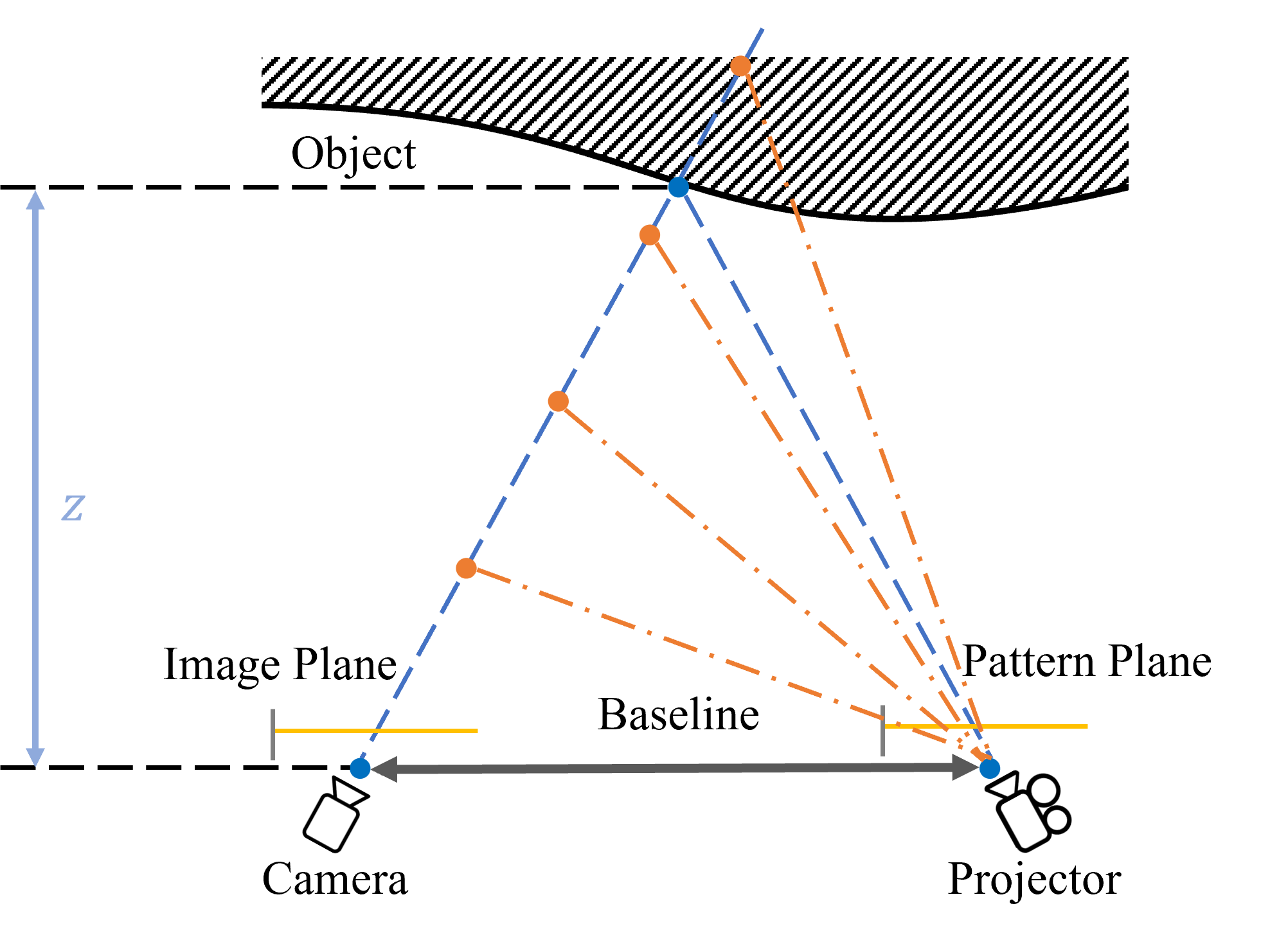}
        \label{fig:ourSL}}
            
	\caption{(a) An illustration of a classic monocular structured light system. Depth is computed by pixel correspondences between the camera and the projector. These correspondences are typically calculated through image matching algorithms. (b) Our monocular structured light system. During the rendering process, we sample points along each camera ray and project them onto the pattern plane to retrieve their corresponding color values.}
	\label{fig:structured light}
\end{figure}

\IEEEPARstart{T}he acquisition of precise depth measurements constitutes a core technical challenge in modern perception pipelines. With the advent of structured-light cameras such as Kinect V2 and Intel RealSense~\cite{keselman2017intel}, structured light systems have become a powerful solution for depth sensing in various applications~\cite{batlle1998recent,cai2020single,salvi2010state}. Typically, a monocular structured light system consists of one camera and one projector with calibrated intrinsic and extrinsic parameters. By projecting randomly or manually designed patterns into 3D space, the system extracts depth information by analyzing the deformation of these patterns in captured images. Classical algorithms in structured light aim to establish robust correspondences across multiple projected patterns. We illustrate this monocular structured light system in Fig.~\ref{fig:monoSL}. The disparity map is computed between the captured image and the projected pattern. Since disparity is inversely proportional to depth, the depth map can be recovered through triangulation, given the known geometric calibration between the camera and the projector. Both disparity maps and depth maps encode scene geometry and are mathematically convertible. Therefore, in this paper, we treat disparity and depth interchangeably without explicit distinction.

The 3D scanning process in structured light systems often requires balancing the trade-off between scan precision and the number of projected patterns or captured images. Increasing the number of captured images can improve the accuracy of correspondence matching. However, more projected patterns also lead to longer acquisition times, which limits the applicability of structured light systems in general scenes, especially those with undesired motion and short exposure times. To address this, researchers have developed techniques to embed richer information within a limited set of patterns~\cite{young2007coded,mirdehghan2018optimal,jia2024adaptivestereo}. These methods use sophisticated patterns that encode temporal or spatial features to mitigate matching uncertainties. By decoding these features from captured images, structured light systems can determine the correspondences needed for accurate depth estimation~\cite{kawasaki2008dynamic,sagawa2009dense,sagawa2011dense,fu2023fast}. 

Despite these advances, designing features that can produce accurate dense depth maps while remaining resilient to environmental influences remains a significant challenge. Traditional structured light depth recovery methods rely on the accuracy of the matching algorithm between projected patterns and captured images. Any errors in the matching process, such as blurring or occlusions, can introduce substantial inaccuracies in the final depth maps. Recent deep learning techniques offer solutions using neural networks to tackle matching uncertainties~\cite{johari2021depthinspace,riegler2019connecting,qiao2022tide}. Most of these methods formulate the depth
estimation process with a convolutional neural network. Schreiberhuber et al.~\cite{schreiberhuber2022gigadepth} formulate depth estimation as a regression problem and address it by discretizing the regression into a series of classification sub-tasks, which were solved using multi-layer perceptrons (MLPs). While these models can generate dense depth maps, they require extensive training datasets, significantly impacting network performance. Given the diversity of devices and pattern configurations, constructing such datasets is challenging in structured light systems.
 
Our work draws inspiration from recent success in using classic voxel grids to explicitly store scene geometries~\cite{garbin2021fastnerf,yu2021plenoctrees,chen2022fast}. Instead of focusing on designing or learning robust matching features, we introduce a novel framework based on volume rendering. Specifically, our approach employs a voxel grid to represent the volume density of the target 3D scene. Through a fully differentiable rendering process, we generate images from the camera's viewpoint, calculating color from projected patterns (as shown in Fig.~\ref{fig:ourSL}), and establish a straight training pipeline with a direct loss function between the captured images and their rendered counterparts. Once training converges, the volume density of the 3D space is obtained, enabling the extraction of both the scene's geometry and the depth map via simple volumetric rendering. The rendering pipeline in our approach is similar to NeRF-based techniques used in view synthesis tasks~\cite{mildenhall2021nerf,yu2021pixelnerf,jin2024reliable}. While these methods achieve impressive results in image synthesis from passive views, they face limitations in geometry recovery due to the need to jointly estimate radiance and geometry fields from captured images~\cite{niemeyer2022regnerf,wang2021neus,yan2021continual,yariv2021volume}. In structured light systems, however, the radiance field is predetermined by the projected patterns, allowing us to focus solely on optimizing the geometry field for high-quality depth estimation.

We leverage constraints from the projected light field to optimize the 3D voxel grid in a monocular structured light setup. This process includes a rendering mechanism that incorporates these projected pattern constraints during ray sampling. Additionally, we introduce a distortion loss to accelerate training and a surface color loss to enhance geometric accuracy. Our experimental results demonstrate that, with as few as six randomly generated binary patterns, our method significantly outperforms existing matching-based techniques in terms of geometric accuracy. Specifically, it reduces the average estimated depth error by approximately 30\% on both synthetic scenes and real-world captured scenes, when compared with previous matching-based methods. 

\begin{figure}[t]
	\centering
	%\fbox{\rule{0pt}{2in} \rule{0.9\linewidth}{0pt}}
	\includegraphics[width=0.9\linewidth]{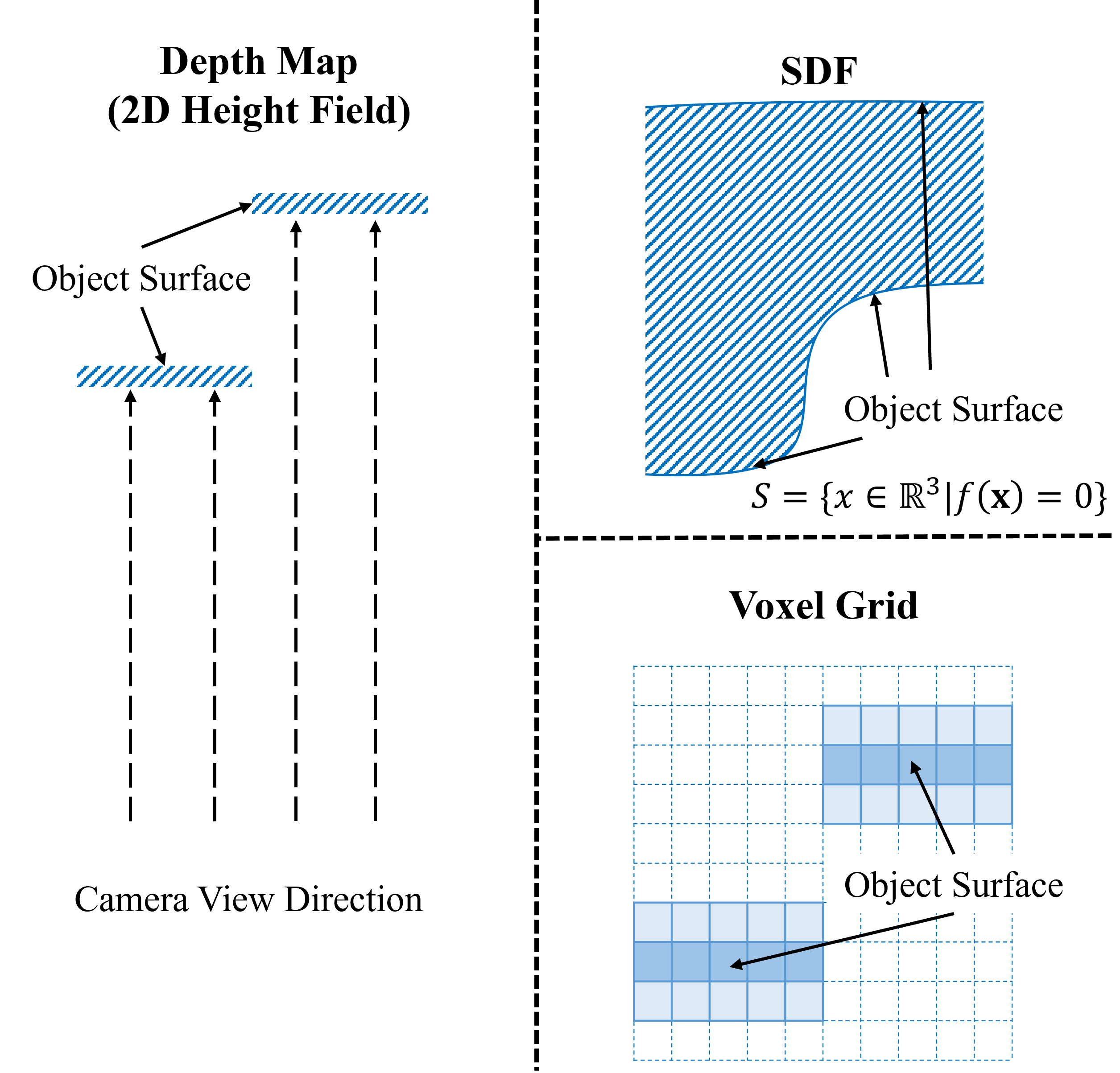}
	
	\caption{Visualization of different geometry representations at object edges of the input scene. The depth map can be treated as a 2D height field with a single value at each 2D location represented by a camera ray. The SDFs generate watertight surfaces, resulting in two surface points along each ray, and leading to difficulties in representing sharp edges. Voxel grids model scene properties at discrete volumetric locations, offering superior flexibility in capturing sharp edges.}
	\label{fig:geovs}
\end{figure}

There are existing methods that utilize volume rendering to address depth recovery tasks in structured light systems. NFSL~\cite{shandilya2023neural} utilizes a neural field to recover the scene’s geometry with a moving camera, extracting correspondences between different input camera views and projected patterns. Our task is quite different from theirs, as our method operates under the monocular structured light system with a single-camera pose. There are also methods considering monocular settings ~\cite{qiao2024depth,mirdehghan2024turbosl}. They train a signed distance function(SDF) as the geometry representation, and apply the marching cubes algorithm~\cite{lorensen1998marching} to extract the depth map. While SDFs can effectively enforce surface smoothness through their implicit continuous formulation, they are not very suitable for this monocular task, as their watertight surfaces inherently result in two surface points along each camera ray, which does not align with the expected characteristics of depth maps (2D height fields). Additionally, employing SDFs in this task leads to difficulties in capturing fine details, especially in regions with sharp depth discontinuities like object boundaries. With only a single viewpoint available, the trained SDF represents the scene's geometry as a single watertight surface, which tends to smooth out geometric discontinuities, making it difficult to recover depth in those areas accurately. In contrast, our explicit voxel grid representation is not constrained by such global modeling, allowing for more localized control, enabling finer precision in areas with significant depth variation. Furthermore, our approach offers a substantial advantage in training efficiency, achieving a nearly three times faster training speed than these methods that rely on implicit geometry representations. We visualize these differences between SDFs and voxel grids in Fig.~\ref{fig:geovs}.

In summary, the main contributions of our work are as follows:
\begin{itemize}[nosep,left=1.2em]
\item We propose a novel matching-free framework for depth estimation in structured light systems that eliminates the need for extensive training datasets or specifically designed patterns typically required in traditional matching-based techniques.

\item Our method incorporates color information from the projected patterns as an existing color field to train the voxel grid, facilitating faster convergence and improving geometry estimation performance.

\item By using voxel grids for geometry representation in our training process, we achieve efficient training speed and good geometry performance, outperforming existing rendering-based methods that utilize implicit representations.

\end{itemize}

\section{Related Work}

\textbf{Temporal-encoding structured light systems.} Temporal-encoding patterns are widely used in structured light systems for static scene reconstruction, as they enable unique decoding at each camera pixel. The plane of light is in a unique location at each time instant, and can be used to recover depth. A number of strategies for coding space with a time sequence of light patterns have been proposed, including binary strips\cite{scharstein2003high}, gray codes~\cite{posdamer1982surface,aliaga2008photogeometric,weinmann2011multi}, grid patterns~\cite{lei2013design}, fringe patterns~\cite{kawasaki2008dynamic,sagawa2011dense,taguchi2012motion}, and phase measurements~\cite{wang20133d,zuo2013high}. Modern structured light systems achieve high accuracy and resilience to noise when sufficient patterns are projected. However, this accuracy declines significantly in few-shot scenarios where only a limited number of patterns can be projected. To address this limitation, researchers have explored enhanced coding strategies, such as using specialized devices~\cite{aliaga2008photogeometric,sundar2022single,weinmann2011multi}, complex pattern designs~\cite{lei2013design,scharstein2003high}, or global optimization methods during post-processing~\cite{kawasaki2008dynamic,koninckx2006real,sagawa2011dense}. Despite these efforts, manually modeling uncertain factors often introduces unstable outliers in the depth map estimation.

\textbf{Learning-based matching in structured light systems.} Recently, learning-based techniques have become popular in many areas and have achieved tremendous success. Deep-learning-based approaches have also been adopted to address the matching problem in structured light systems, as demonstrated in studies like ~\cite{fanello2016hyperdepth,fanello2017ultrastereo}. Some of these methods formulate the depth estimation process with a convolutional
network. Some other works employ deep networks to directly predict disparity from image-pattern pairs within these systems\cite{johari2021depthinspace,riegler2019connecting,zhang2018activestereonet}. Schreiberhuber et al.~\cite{schreiberhuber2022gigadepth} formulate the depth estimation task as a regression problem. However, these approaches require extensive training datasets, and the scarcity of publicly available benchmarks tailored to the unique characteristics of structured light—such as specific pattern designs and parameter configurations—presents a significant challenge. Besides, these learning-based methods often encounter domain shift issues when applied to real scene settings. 

\textbf{Voxel grid representation for 3D geometry.} Voxel grids model 3D objects by dividing space into a regular array of volumetric units, or voxels, each storing data like color, density, or material properties. This approach is advantageous for capturing complex geometries and internal structures, especially when surface-based methods, such as meshes, fall short. Noteworthy applications of voxel grids in deep learning include VoxNet by Maturana and Scherer~\cite{maturana2015voxnet}, a 3D CNN that operates on voxel grids for object recognition. Further advancing this integration, VoxGRAF by Schwarz et al.~\cite{schwarz2022voxgraf} introduces a sparse voxel grid framework for 3D-aware image synthesis, efficiently rendering views with 3D consistency and visual fidelity by combining sparse grids with 3D convolutions, progressive growing, and free-space pruning. DVGO by Cheng Sun et al.~\cite{sun2022direct} applies voxel grids to model 3D geometry in a neural rendering pipeline~\cite{mildenhall2020nerf}, which achieves NeRF-comparable quality and converges rapidly from scratch. In this work, we also adopt a voxel grid structure to develop a tailored training framework for high-quality depth reconstruction within structured light systems.

\section{Methodology}

Our approach focuses on monocular structured light systems, consisting of a single camera and a projector to capture images for depth estimation. Given a set of patterns $\left\{\mathbf{P}_{i}\right\}_{i=1,2, \ldots N}$, the projector sequentially projects them onto the scene, while the camera captures images $\left\{\mathbf{I}_{i}\right\}_{i=1,2, \ldots N}$. Here, $N$ represents the number of projected patterns. With the known intrinsic and extrinsic parameters of the camera and the projector, we aim to reconstruct the depth map $\mathbf{D}$ from the camera viewpoint. As it is under monocular setting, we can simply set the extrinsic matrix of the camera to the identity matrix.

\begin{figure*}[t]
	\centering
	%\fbox{\rule{0pt}{3in} \rule{0.9\linewidth}{0pt}}
	\includegraphics[width=0.95\linewidth]{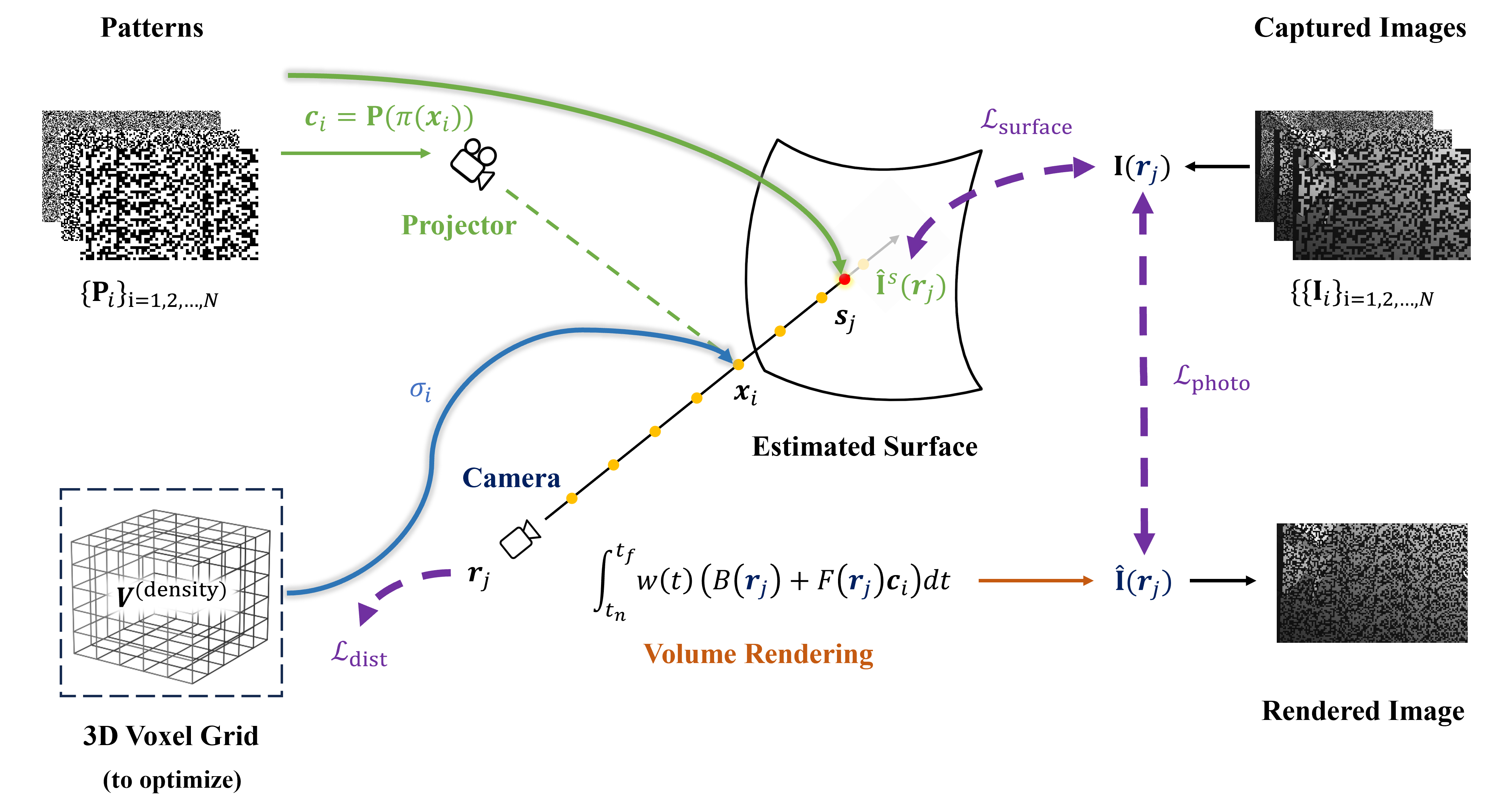}
	
	\caption{Our pipeline for matching-free depth recovery in structured light systems. Sampled rays are rendered through a volume rendering process using voxel grid density. Each sampled point along a ray is projected into the projector space to retrieve its corresponding color value from the structured light pattern. To supervise the voxel grid optimization, both the rendered color of the entire ray and the color of its estimated surface point are compared against the captured image.}
	\label{fig:pipeline}
\end{figure*}

Traditional matching-based methods attempt to define a function that uses estimated point-by-point correspondences to directly compute the depth map. In contrast, we introduce a novel matching-free framework, as shown in Fig.~\ref{fig:pipeline}. We first construct a density voxel grid to store the geometric information of the input scene (see Section~\ref{sec:densityvoxelgrid}). Next, we employ a differentiable volume rendering process using the voxel grid and the projected patterns to generate images under the camera view (see Section~\ref{sec:rendering}). During the optimization process, we utilize several loss functions to compare the rendered images with the captured ones and to encourage voxel densities to be compact and sparse (see Section~\ref{sec:losses}). Finally, we introduce the NDC parameterization, which reallocates the voxel grid's density to better align with the geometry of perspective projection (see Section~\ref{sec:ndc_p}).

\subsection{Density Voxel Grid}
\label{sec:densityvoxelgrid}

A voxel-grid representation explicitly encodes relevant scene modalities (e.g. density, color, or feature) within each grid cell. This structured approach allows for efficient interpolation-based queries at any 3D position, facilitating rapid access to detailed spatial information:
\begin{equation}
\operatorname{interp}(\boldsymbol{x}, \boldsymbol{V}):\left(\mathbb{R}^{3}, \mathbb{R}^{C \times N_{x} \times N_{y} \times N_{z}}\right), \rightarrow \mathbb{R}^{C},
\label{eq:voxelgrid}
\end{equation}
where $\boldsymbol{x}$ denotes the queried 3D point, $\boldsymbol{V}$ represents the voxel grid, $C$ is the modality dimension, and $N_x , N_y , N_z$ is total number of voxels on each dimension. We use trilinear interpolation in our method.

Density voxel grid $\boldsymbol{V}^{\text {(density) }}$ is a special case with $C = 1$, which stores the density values for volume rendering (see Eq.\eqref{eq:rayrender}). To optimize voxel density directly, we use $\ddot{\sigma} \in \mathbb{R}$ to denote the raw density stored by the voxel grid and use the shifted $\operatorname{softplus}$ from Mip-NeRF~\cite{barron2021mip} to apply the density activation:
\begin{equation}
	\sigma=\operatorname{softplus}(\ddot{\sigma})=\log (1+\exp (\ddot{\sigma}+b)).
	\label{eq:activation}
\end{equation}
All grid values in $\boldsymbol{V}^{\text {(density) }}$ are initially set to zero. Employing the $\operatorname{softplus}$ function instead of $\operatorname{ReLU}$ is essential for optimizing voxel density, as $\operatorname{ReLU}$ may irreversibly set it to zero when a voxel value is falsely set to negative.

To ensure that all sampled points on rays are visible to the camera at the start, we initialize the accumulated transmittance $T_{i}\approx1$ by setting the shift $b$ to 
\begin{equation}
	b=\log \left(\left(1-\alpha^{(\text {init })}\right)^{-\frac{1}{\delta}}-1\right),
	\label{eq:bias}
\end{equation}
where $\alpha^{(\text {init })}$ serves as a hyperparameter, and $\delta$ is the step size. 

\subsection{Rendering with Projected Patterns}
\label{sec:rendering}

For one of the captured images $\mathbf{I}_{j}$, we define its corresponding rendered image as $\widehat{\mathbf{I}}_{j}$. To render the color of a pixel $\widehat{\mathbf{I}}_{j}(\boldsymbol{r})$, we cast the ray $\boldsymbol r$ from the camera center through the pixel. $K$ points are then sampled on $\boldsymbol r$ between the predefined near and far planes. The $K$ ordered sampled points are then used to query for their densities and colors $\left\{\left(\sigma_{i}, \boldsymbol{c}_{i}\right)\right\}_{i=1}^{K}$. Finally, the $K$ queried results are accumulated into a single color with the volume rendering equation~\cite{mildenhall2021nerf}:

\begin{equation}
	\begin{aligned}
		\widehat{\mathbf{I}}_{j}(\boldsymbol{r}) & =\left(\sum_{i=1}^{K} T_{i} \alpha_{i} \boldsymbol{c}_{i}\right) \\
		\alpha_{i} & =1-\exp \left(-\sigma_{i} \delta_{i}\right)\\ T_{i} & =\prod_{j=1}^{i-1}\left(1-\alpha_{j}\right),
	\end{aligned}
	\label{eq:rayrender}
\end{equation}
where $\alpha_{i}$ is the probability of termination at point $\boldsymbol{x}_i$; $T_{i}$ is the accumulated transmittance from the near plane to point $i$, and $\delta_{i}$ is the distance to the adjacent sampled point.

We use the post-activation strategy from~\cite{sun2022direct} to calculate the density of the sampled point $\boldsymbol{x}_{i}$. That means we first use trilinear interpolation to get the raw density value:
\begin{equation}
	\ddot{\sigma_{i}}=\operatorname{interp}\left(\boldsymbol{x}_{i}, \boldsymbol{V}^{(\text {density)}}\right)
	\label{eq:rawdensity}
\end{equation}
We then employ the softplus function (Eq.\eqref{eq:activation}) to get $\sigma_{i}$, and finally use it to calculate rendered results. DVGO~\cite{sun2022direct} has shown that the post-activation strategy can produce a sharp linear boundary. 

In our approach, the color of the sampled point $\boldsymbol{x}_{i}$ is not estimated from the voxel grid or the training schedule. Instead, we utilize prior knowledge from our projected patterns to determine the point color using a re-projection function $\pi$. Specifically, the color $\boldsymbol{c}_{i}$ can be directly calculated from the projected pattern $\mathbf{P}_{j}$ through:
\begin{equation}
	\boldsymbol{c}_{i}=B(\boldsymbol{r})+F(\boldsymbol{r}) \mathbf{P}_{j}(\pi(\boldsymbol{x}_{i})),
	\label{eq:projection}
\end{equation}
where $\boldsymbol{r}$ is the sampled ray, $\mathbf{P}_{j}(\pi(\boldsymbol{x}_{i}))$ symbolizes the projected color calculated through re-projection operation based on the intersection between projected light and sampled ray (note that $\mathbf{P}(\cdot)$ query pixel color on the pattern via its pixel coordinates). $B(\boldsymbol{r})$ and $F(\boldsymbol{r})$ stand for the background light level and fringe contrast, respectively. They are calculated from the captured images:
\begin{equation}
	\begin{array}{l}
		B(\boldsymbol{r})=\min \left(\left\{\mathbf{I}_{j}(\boldsymbol{r})\right\}_{j=1,2, \ldots N}\right) \\
		F(\boldsymbol{r})=\max \left(\left\{\mathbf{I}_{j}(\boldsymbol{r})\right\}_{j=1,2, \ldots N}\right)-B(\boldsymbol{r}).
	\end{array}
	\label{eq:B&F}
\end{equation}

It is important to highlight that these two parameters inherently enable the masking of occluded regions, as both the fringe contrast and the background light level tend to approach zero in these areas.

\subsection{Loss Functions}
\label{sec:losses}

During the training process, we pick a batch of $M$ pixels from each captured image, and sample $K$ points on each corresponding ray. We then calculate the color of these rays to generate the rendered color $\widehat{\mathbf{I}}_{j}(\boldsymbol{r})$ for each image $j$. We first minimize a per-pixel loss function that quantifies the difference between $\widehat{\mathbf{I}}(\boldsymbol{r})$ and the color $\mathbf{I}(\boldsymbol{r})$ from the captured image. The photometric MSE is defined as 
\begin{equation}
	\mathcal{L}_{\text {photo }}=\frac{1}{MN} \sum_{j=1}^{N}\sum_{i=1}^{M}\|\widehat{\mathbf{I}}_{j}(\boldsymbol{r}_{i})-\mathbf{I}_{j}(\boldsymbol{r}_{i})\|_{2}^{2}.
	\label{eq:Loss_mse}
\end{equation}

\begin{figure*}[t]
	\centering
	%\fbox{\rule{0pt}{2in} \rule{0.9\linewidth}{0pt}}
	\includegraphics[width=0.95\linewidth]{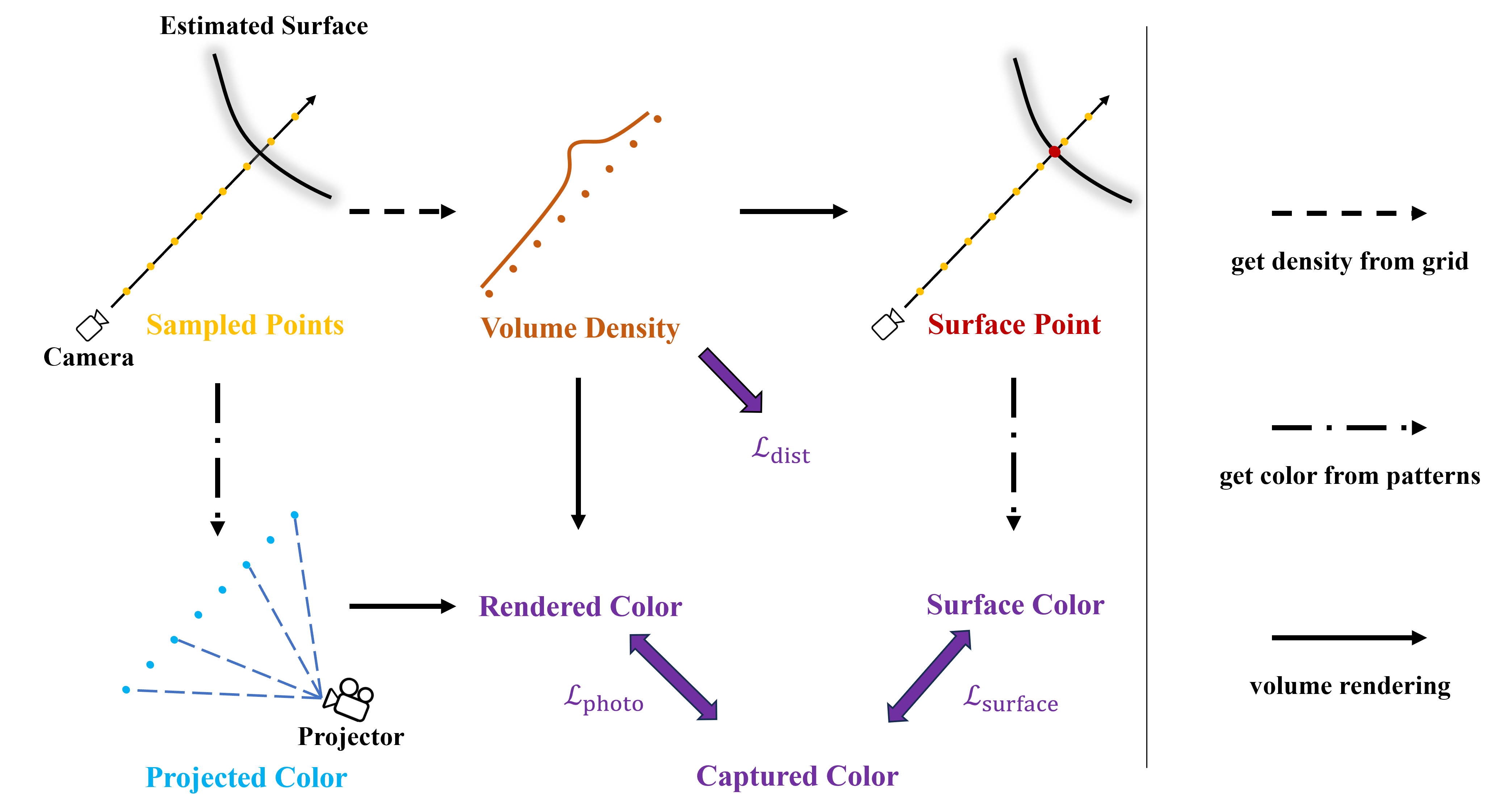}
	
	\caption{Visualization of our photometric loss $	\mathcal{L}_{\text {photo }}$, distortion loss $\mathcal{L}_{\text {dist }}$, and surface color loss $\mathcal{L}_{\text {surface}}$. The photometric loss encourages the rendered colors to match the captured image, ensuring overall appearance consistency. In contrast, both the distortion loss and the surface color loss promote a single dominant peak in the volume density distribution along each ray, thereby improving surface localization and reducing ambiguity in depth estimation.}
	\label{fig:lsc}
\end{figure*}

In our scenario, the volumetric density along each sampled light ray should be singular peaked, as each ray from the camera should intersect only once with the object's surface. Besides, due to the lack of camera views in our task, there may be multiple peaks along the projecting ray to produce one precise color, resulting in lots of floaters in the reconstructed geometry. Thus, we apply a distortion loss proposed by Mip-NeRF 360~\cite{barron2022mip}. For a ray with $K$ sampled points, this loss is defined as
\begin{equation}
	\begin{array}{r}
		\mathcal{L}_{\text {dist }}(s, w)=\sum\limits_{i=0}^{K-1}\sum\limits_{j=0}^{K-1}w_{i} w_{j}\left|\frac{s_{i}+s_{i+1}}{2}-\frac{s_{j}+s_{j+1}}{2}\right| 
		\\+\frac{1}{3} \sum\limits_{i=0}^{K-1} w_{i}^{2}\left(s_{i+1}-s_{i}\right),
	\end{array}
	\label{eq:distortion}
\end{equation}
where $(s_{i+1}-s_{i})$ is the length, $(s_{i}+s_{i+1})/2$ is the midpoint of the $i$-th query interval and $w_{i}=T_{i}\alpha_{i}$. The first term minimizes the weighted distances between all pairs of interval midpoints, and the second term minimizes the weighted size of each individual interval. As a result, we force the trained geometry to fit our task. The total distortion loss is
\begin{equation}
	\mathcal{L}_{\text {dist-total }}=\frac{1}{MN}\sum_{j}\sum_{i}\mathcal{L}_{\text {dist }}.
	\label{eq:totaldist}
\end{equation}

We additionally introduce another color loss to deal with those artifacts. We first compute the rendered surface point of a ray $\mathbf{r}_{l}$ using a volume rendering equation similar to Eq.\eqref{eq:rayrender} :
\begin{equation}
\boldsymbol{s}_{l} =\left(\sum_{i=1}^{K} T_{i} \alpha_{i} \boldsymbol{x}_{i}\right).
	\label{eq:depthrender}
\end{equation}
Then Eq.\eqref{eq:projection} is applied to compute the pixel color of $\boldsymbol{s}_{l}$ as
\begin{equation}
	\widehat{\mathbf{I}}_{j}^{s}(\boldsymbol{r}_{l})=B(\boldsymbol{r}_{l})+F(\boldsymbol{r}_{l}) \mathbf{P}_{j}(\pi(\boldsymbol{s}_{l})).
	\label{eq:surfacecolor}
\end{equation}

We formulate the surface color loss as
\begin{equation}
	\mathcal{L}_{\text {surface}}=\frac{1}{MN} \sum_{j=1}^{N}\sum_{i=1}^{M}\|\widehat{\mathbf{I}}_{j}^{s}(\boldsymbol{r}_{i})-\mathbf{I}_{j}(\boldsymbol{r}_{i})\|_{2}^{2}.
	\label{eq:lsc}
\end{equation}
The surface color loss is similar to photometric constraints, which are applied in matching-based techniques~\cite{garg2016unsupervised,zhang2018activestereonet}, where colors are warped between images using pixel depth information. It can enhance geometry quality and overall performance, especially in few-shot scenarios~\cite{darmon2022improving,niemeyer2022regnerf,xiangli2022bungeenerf}. Our approach prioritizes geometric performance over photorealism. Therefore, enforcing density constraints helps the voxel grid generate more accurate 3D shapes without introducing rendering ambiguities. We provide a schema to illustrate the difference between our losses in Fig.~\ref{fig:lsc}.

The whole loss function is defined as
\begin{equation}
	\mathcal{L}=\mathcal{L}_{\text {photo }}+\lambda_{d}\mathcal{L}_{\text {dist-total }}+\lambda_{s}\mathcal{L}_{\text {surface}}.
	\label{eq:fullloss}
\end{equation}
Once the training process is finished, we extract the full depth map $\mathbf{D}$ of the input scene using Eq.\eqref{eq:depthrender}.

\subsection{NDC Parameterization}
\label{sec:ndc_p}
\begin{figure}[tbp]
	\centering
	%\fbox{\rule{0pt}{2in} \rule{0.9\linewidth}{0pt}}
	\includegraphics[width=0.9\linewidth]{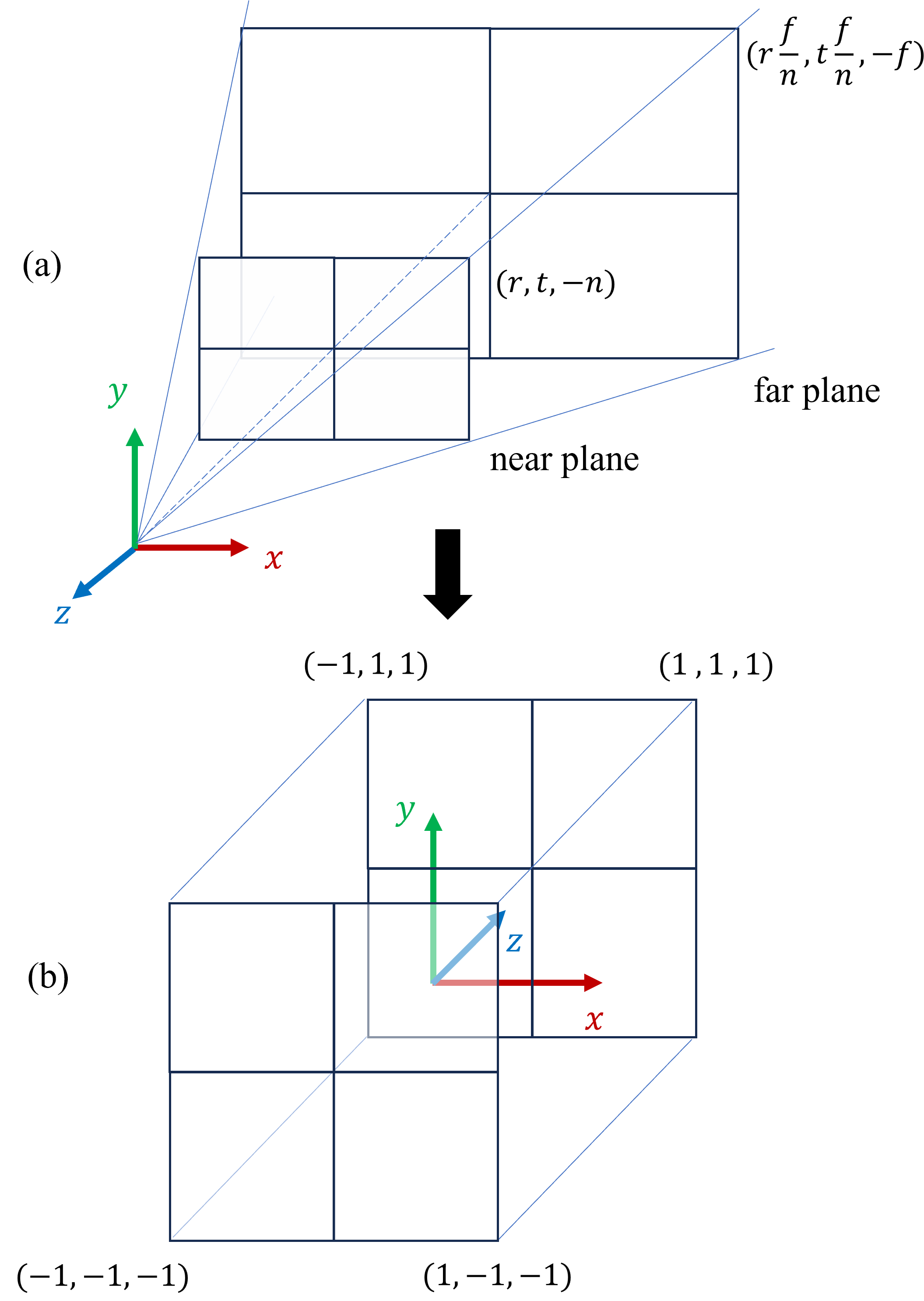}
	
	\caption{Transformation from camera frustum in world space to NDC space. (a) world coordinates. (b) normalized device coordinates. We define the near and far planes as the depth boundaries for this transformation.}
	\label{fig:ndc_vis}
\end{figure}

Our task focuses on front-facing scenes, as there's only one camera in a monocular structured light system. Inspired by the original NeRF~\cite{mildenhall2021nerf}, we define our voxel grid in normalized device coordinates(NDC) space. The transformation from the camera frustum to the NDC space in our process is illustrated in Fig.~\ref{fig:ndc_vis}. The standard 3D perspective projection matrix for homogeneous coordinates is given by:
\begin{equation}
	M=\left(\begin{array}{cccc}
\frac{n}{r} & 0 & 0 & 0 \\
0 & \frac{n}{t} & 0 & 0 \\
0 & 0 & \frac{-(f+n)}{f-n} & \frac{-2 f n}{f-n} \\
0 & 0 & -1 & 0
\end{array}\right)
	\label{eq:ndcmatrix}
\end{equation}
where $n$ and $f$ represent the near and far clipping planes, and $r$ and $t$ are the right and top bounds of the frustum at the near clipping plane. Given a homogeneous point $(x, y, z, 1)^T$, we apply the transformation by left-multiplying the point by the matrix $M$ and then dividing out the fourth coordinate to get the projected point
\begin{equation}
(-\frac{nx}{rz},-\frac{ny}{tz},\frac{(f+n)}{f-n}+\frac{2 f n}{(f-n)z})^T
\label{eq:ndcpro2}
\end{equation}
The projected point is now in normalized device coordinate (NDC) space, where the original viewing frustum is mapped to the cube $[-1, 1]^3$. By mapping the top-right pixel on the image plane to the top-right corner of the near plane, we obtain:
\begin{equation}
    \frac{f_x}{c_x}\frac{r}{n}=1,\quad\quad \frac{f_y}{c_y}\frac{t}{n}=1
\label{eq:ndcpro3}
\end{equation}
Where $n$ represents the chosen near clipping plane, $f_x,f_y$ and $c_x,c_y$ are focal lengths and principal points of the camera. Therefore, $r$ and $t$ can be computed using the camera parameters:
\begin{equation}
    r=\frac{n c_x}{f_x},\quad\quad t=\frac{n c_y}{f_y}
\label{eq:ndcpro4}
\end{equation}

We further set $f \to \infty$ in our approach, deriving the relationship between world coordinate $(x,y,z)^T$ and its corresponding NDC $(x_*,y_*,z_*)^T$
\begin{equation}
\begin{split}
    & x = 2n\frac{c_x}{f_x}\frac{x_*}{1-z_*} \\
    & y= 2n\frac{c_y}{f_y}\frac{y_*}{1-z_*} \\
    & z=\frac{2n}{z_*-1} 
\end{split}
\label{eq:ndcpro5}
\end{equation} 
Here $z\in (-\infty,-n]$ and $z_*\in [-1,1)$ as the camera is looking in the $-z$ direction.

By warping an infinitely deep camera frustum into a bounded cube, where distance along the z-axis corresponds to disparity (inverse distance), NDC efficiently reallocates the voxel grid's density in a way that aligns with the geometry of perspective projection. 

It is important to note that NDC is particularly well-suited for our task. When we uniformly sample points along a ray in NDC space, these samples are uniformly spaced in disparity. According to our projection matrix $\mathbf{P}$, the offset between two sampled points along a camera ray in pattern coordinates is also proportional to disparity. As a result, the sampled points are evenly distributed across the patterns.

\section{Experiments}

\subsection{Datasets}

\begingroup
\setlength{\tabcolsep}{6pt} % Default value: 6pt
\renewcommand{\arraystretch}{1} % Default value: 1
\begin{table}[]
\centering
\caption{\centering Camera and projector parameters used in real-world scenes}
\label{tab:cam_para}
\begin{tabular}{c|ccccc}
\toprule
Device    & $f_x$   & $f_y$   & $c_x$  & $c_y$  & \multicolumn{1}{l}{Baseline} \\ 
\midrule
Camera    & 1181.76 & 1179.92 & 639.50 & 511.50 & \multirow{2}{*}{209.39} \\
Projector & 2013.30 & 2016.43 & 699.16 & 755.26 & \\ 
\bottomrule
\end{tabular}
\end{table}
\endgroup
 
We use patterns as color constraints for training the voxel grid to represent 3D scene geometry, eliminating the need for explicit matching techniques that are applied to pattern sets and captured images. Thus, we simply use a set of randomly generated 2D binary patterns. The projector space is divided into uniform squares, with each square randomly assigned either black or white. To generate our pattern sets, we use unit squares of varying sizes. We select pattern lengths of 20, 10, and 5 pixels to generate a total of 6 patterns (two patterns per scale) for experiments. 

To evaluate our approach and compare it with existing methods, we first assess the accuracy of depth estimation on synthetic scenes. These scenes are rendered using the tool provided by CTD~\cite{riegler2019connecting}, following the same experimental setup as CTD and using objects from the ShapeNet dataset~\cite{chang2015shapenet}. A total of 50 different scenes are rendered to create our synthetic dataset. Additionally, we test our method on real scenes provided by SL-SDF~\cite{qiao2024depth}, which includes six scenes with estimated ground truth depths. Device parameters of these scenes are shown in Table~\ref{tab:cam_para}.

\begingroup
\setlength{\tabcolsep}{10pt} % Default value: 6pt
\renewcommand{\arraystretch}{1.2} % Default value: 1
\begin{table*}[htb]%
\centering
\caption{Pattern Types, Number of Patterns, and Pretraining Requirements of Different Methods}
\label{tab:patterns}
\begin{tabular}{c|cccccccc}
\toprule
                        & N-PMP & H-PMP & CGC & GC &  CTD  & GigaDepth  &SL-SDF & Ours \\ 
\midrule
Binary Patterns(1D)     & -     & -     & 4   & 9  & -      & -    & -    & -    \\
Sinusoidal Patterns(1D) & 6     & 6     & 3   & -  & -      & -    & -    & -    \\
Binary Patterns(2D)     & -     & -     & -   & -  & -      & -    & 6    & 6    \\
Speckle Patterns(2D)    & -     & -     & -   & -  & 1      & 1    & -    & -    \\
\midrule
Pretraining Datasets   & -     & -     & -   & -  & required   & required & -    & -    \\
\bottomrule
\end{tabular}
\end{table*}%
\endgroup

\subsection{Implementation Details}

We use consistent hyperparameters across all scenes. We divide the xy-plane of the NDC cube into $256\times 256$ and discretize the length along the z-axis into $256$ different disparity values, resulting in an expected voxel count of $M=256^3$. We set $\alpha^{(\text {init })}=10^{-2}$ and the point sampling step size is chosen as half of the voxel size. For training the voxel grid, we use $\lambda_d=0.01$ with $\lambda_s=0$ for the first 3000 iterations, during which we sample 8192 rays per iteration. The exclusion of surface color loss during this phase is a strategic choice to avoid potential issues with local minima. After the initial phase, we set $\lambda_s=1$ and continue training for an additional 29,000 iterations. The entire training process takes approximately 5 minutes on a single NVIDIA GTX 4060.

\subsection{Comparisons}
\begin{figure*}[t]%
	\centering
	%\fbox{\rule{0pt}{2in} \rule{0.9\linewidth}{0pt}}
	\includegraphics[width=0.9\linewidth]{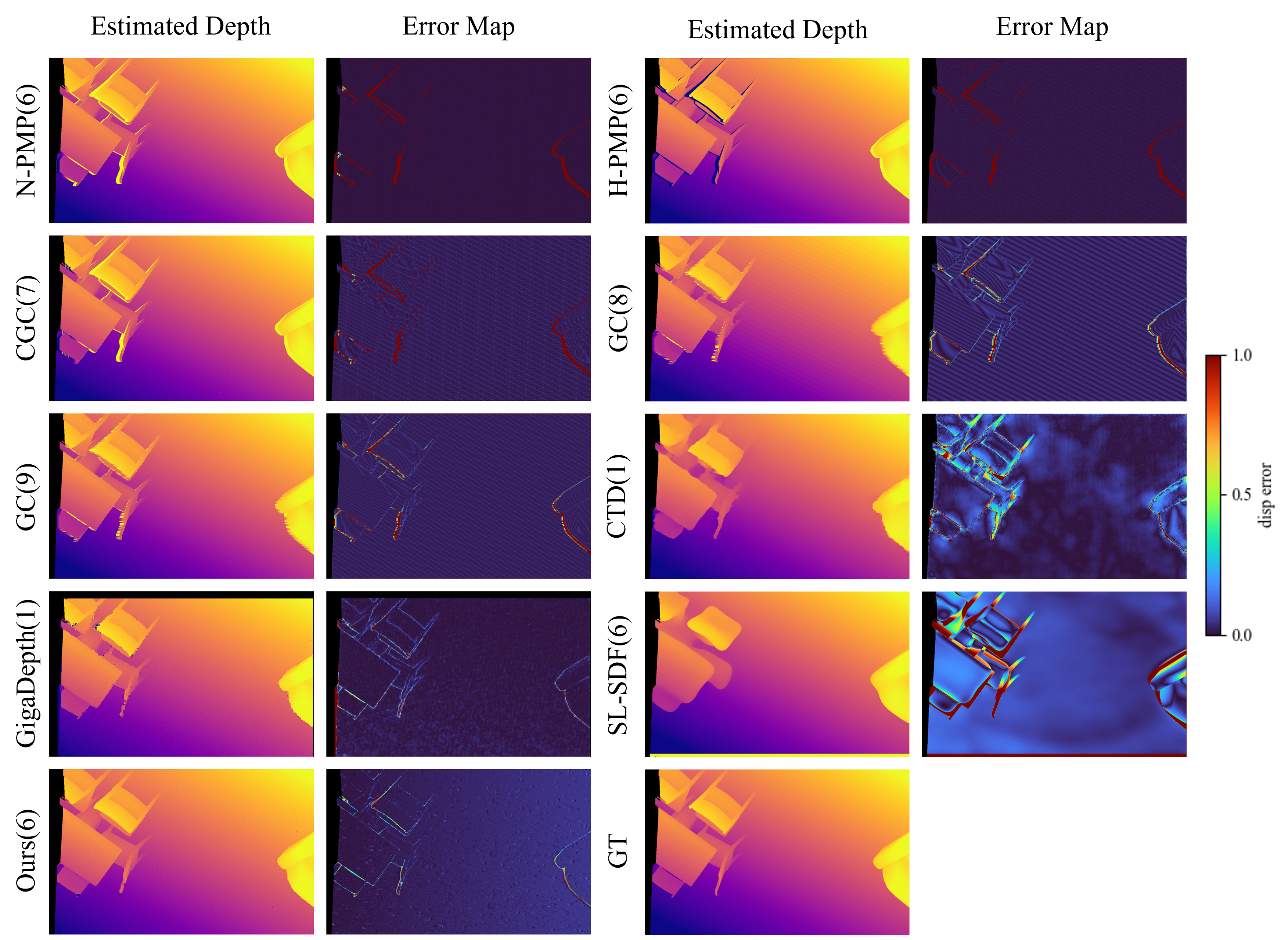}
	\caption{Visualization of estimated depth maps and corresponding error maps from different methods on synthetic scenes.}
	\label{fig:resultvis}
\end{figure*}%
\begin{figure*}[t]%
	\centering
	%\fbox{\rule{0pt}{2in} \rule{0.9\linewidth}{0pt}}
	\includegraphics[width=0.9\linewidth]{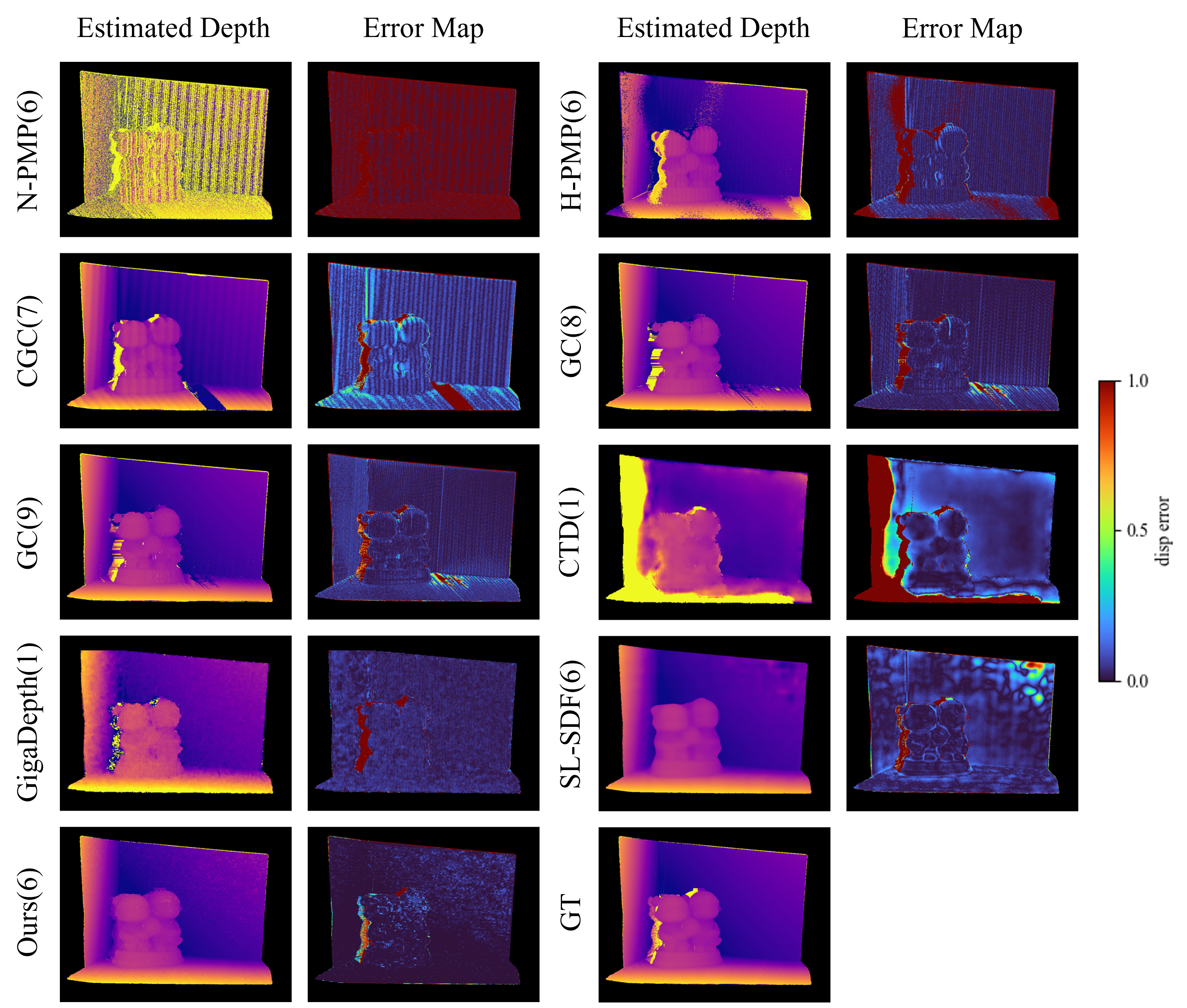}
	\caption{Visualization of estimated depth maps and corresponding error maps from different methods on real scenes.}
	\label{fig:resultvis_real}
\end{figure*}%

\begingroup
\setlength{\tabcolsep}{10pt} % Default value: 6pt
\renewcommand{\arraystretch}{1.2} % Default value: 1
\begin{table*}[]
\caption{Quantitative Comparison of Depth Recovery Performance Across Synthetic and Real-World Scenes}
	\label{tab:result}
	\centering
	\begin{tabular}{c |c c c c|c c c c}
		\toprule
		\multicolumn{1}{c|}{\multirow{2}{*}{Method}} & \multicolumn{4}{c|}{synthetic scenes}                           & \multicolumn{4}{c}{real scenes}                                \\ 
		\multicolumn{1}{c|}{\quad}                        & MAE(mm)      & $O(0.1)$      & $O(0.5)$      & $O(1)$       & MAE(mm)     & $O(0.1)$      & $O(0.5)$      & $O(1)$      \\
		\midrule
		N-PMP(6)~\cite{zuo2013high}                                   & 29.431       & 2.05       & 1.85       & 1.71              & 513.855        & 68.99         & 67.13         & 62.22       \\
		H-PMP(6)~\cite{wang20133d}                                    & 30.787       & 2.16       & 1.09       & 0.61       
        & 56.635         & 14.52         & 3.11          & 2.12      \\
		CGC(7)~\cite{wu2020high}                                      & 33.445       & 2.04       & 1.72       & 1.59              & 38.849         & 6.00          & 3.90          & 1.52        \\
		GC(8)~\cite{aliaga2008photogeometric,weinmann2011multi}                                      & 22.539       & 1.36       & 1.03       & 0.96         & 18.941         & 4.92          & 3.17          & 2.20      \\
		GC(9)~\cite{aliaga2008photogeometric,weinmann2011multi}                                       & 19.087       & 1.02       & 0.62       & 0.57              & 16.171         & 4.30          & 2.19          & 1.23       \\
            CTD(1)~\cite{riegler2019connecting}     & 15.613     &  0.72     &  0.10     &  0.10   & 296.560     &  18.00     &  16.24       & 16.01 \\
            GigaDepth~\cite{schreiberhuber2022gigadepth} & \colorbox[HTML]{FFCE93}{14.176}      & \colorbox[HTML]{FFCE93}{0.10}     &  \colorbox[HTML]{FFCE93}{0.06}     &  \colorbox[HTML]{FFCE93}{0.06} & 18.629     &  \colorbox[HTML]{FFCE93}{2.98}    &   1.47     &  0.73 \\ 
            SL-SDF(6)~\cite{qiao2024depth}                                   & 90.070       & 5.51       & 2.14       & 2.14        & \colorbox[HTML]{FFCE93}{9.376}          & 3.14          & \colorbox[HTML]{FD6864}{\textbf{0.82}} & \colorbox[HTML]{FD6864}{\textbf{0.45}}  \\
		Ours                                        & \colorbox[HTML]{FD6864}{\textbf{13.767}} & \colorbox[HTML]{FD6864}{\textbf{0.08}} & \colorbox[HTML]{FD6864}{\textbf{0.06}} & \colorbox[HTML]{FD6864}{\textbf{0.06}}   & \colorbox[HTML]{FD6864}{\textbf{9.153}} & \colorbox[HTML]{FD6864}{\textbf{2.87}} & \colorbox[HTML]{FFCE93}{0.92}          & \colorbox[HTML]{FFCE93}{0.47}         \\
		\bottomrule
	\end{tabular}
\end{table*}
\endgroup

We first evaluate our method against four classic decoding-based structured light techniques. Specifically, we compare with Numerical Phase Measurement Profilometry (N-PMP)~\cite{zuo2013high}, Hierarchical Phase Measurement Profilometry (H-PMP)~\cite{wang20133d}, Binary Gray Code (GC) with interpolation between fringes~\cite{aliaga2008photogeometric,weinmann2011multi}, and Complementary Gray Code (CGC)~\cite{wu2020high}. N-PMP and H-PMP require six projected patterns, and CGC requires seven. The GC method can take a different number of patterns for calculation, here we choose eight and nine for comparison. We then compare our method with two learning-based methods, Connecting the Dots(CTD)~\cite{riegler2019connecting} and GigaDepth~\cite{schreiberhuber2022gigadepth}. CTD formulates depth estimation as a convolutional neural network task, while GigaDepth models it as a regression problem and further decomposes the regression into smaller classification sub-tasks using multi-layer perceptrons (MLPs). Both methods require pretraining on hundreds of synthetically generated scenes to achieve satisfactory performance. Besides, We also compare our method with SL-SDF~\cite{qiao2024depth}, which is a matching-free approach. ~\cite{qiao2024depth} builds a neural signed distance field to represent 3D geometry, and applys a NeuS-based~\cite{wang2021neus} differentiable rendering scheme during training phase. The depth map is then generated through the marching cubes algorithm~\cite{lorensen1998marching} and re-projection from the trained SDF.

\begin{figure*}
	\centering
	%\fbox{\rule{0pt}{2in} \rule{0.9\linewidth}{0pt}}
	\includegraphics[width=0.9\linewidth]{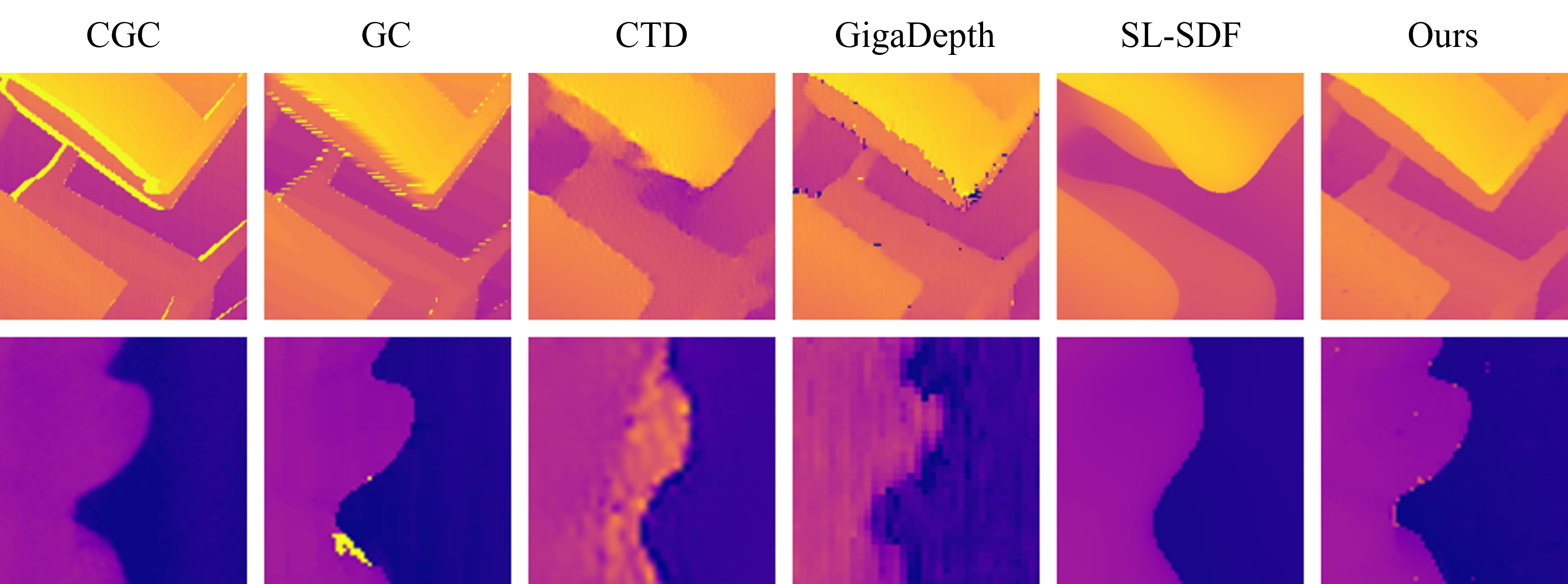}
	\caption{Visualization of the differences in estimated depth maps at object boundaries between our method and prior approaches.}
	\label{fig:sharp edges}
\end{figure*}

We illustrate the types and numbers of patterns used by each method with the requirements of pretraining on generated datasets in Table~\ref{tab:patterns}. To ensure a fair comparison with SL-SDF, we use the same set of 2D projected binary patterns as employed in their approach. We use both the mean absolute error and the outlier metric to evaluate the recovered depth maps of each method. When calculating MAE for classic decoding-based methods, we set the depth value of outliers to the mean value of the estimated depth map to minimize their influence. The definition of the percentage of outliers $o(t)$ is from CTD~\cite{riegler2019connecting}, which is the percentage of pixels where the difference between the estimated and ground-truth disparities (inverse depths) is greater than a certain threshold $t$. We summarize our experimental results and show them in Table~\ref{tab:result}. Results shown in the table represent the average depth error and percentage of outliers across all scenes in the dataset(50 synthetic scenes and six real scenes). Additionally, we visualize the recovered depth maps and error maps by different methods in Fig.~\ref{fig:resultvis} and Fig.~\ref{fig:resultvis_real}. GT stands for the ground truth. Here, we use disparities for better visualization. We also show edge details from different methods in Fig.~\ref{fig:sharp edges} for further comparisons. 

The N-PMP, H-PMP, and CGC methods employ phase-shifting encoding, which represents pixel coordinates using sets of sinusoidal patterns. However, due to their periodic nature, these patterns introduce phase ambiguities during decoding. This issue is typically mitigated by projecting additional patterns at varying spatial frequencies. While phase-shifting techniques can achieve high accuracy with a sufficient number of patterns, they become less reliable when the number of patterns is limited, especially in captured real scenes.

N-PMP utilizes two sets of phase patterns with shorter wavelengths, which increases its sensitivity to phase decoding errors. H-PMP supplements traditional phase patterns with an additional pattern set having a wavelength equal to the image width, but remains vulnerable to shading effects and intensity variation. CGC incorporates binary Gray-code patterns to resolve ambiguities; however, the long wavelengths of these binary patterns limit their decoding precision. GC, which relies solely on binary Gray-code patterns, is prone to errors between adjacent binary fringes due to limited spatial resolution. 

CTD and GigaDepth perform well on synthetic datasets, benefiting from carefully designed training scenes with ideal lighting and noise-free conditions. However, their effectiveness diminishes when applied to real captured scenes, primarily due to a significant domain gap between synthetic and real-world data. Moreover, both methods struggle to preserve fine geometric details at object boundaries. CTD’s reliance on convolutional architectures may limit its ability to handle high-frequency variations, while GigaDepth’s discretization of the regression task can lead to quantization artifacts near depth discontinuities. As a result, their depth estimations at object edges often appear overly smoothed or inaccurate, which is particularly problematic in applications requiring precise surface reconstruction.

Both SL-SDF~\cite{qiao2024depth} and our approach employ 2D binary patterns and eliminate the need for explicit pattern matching. Experimental results indicate that both methods can generate smooth surface reconstructions. However, SL-SDF exhibits degraded performance in certain synthetic scenes, particularly those containing sharp edges or intricate geometric structures. This limitation arises from its use of signed distance fields (SDFs) as the underlying geometry representation.

While SDFs inherently promote surface smoothness due to their continuous and differentiable formulation, they are less effective at preserving fine-scale geometric discontinuities. This drawback is especially evident in structured light scenarios, where high-frequency textures and rich color cues are largely absent. Consequently, sharp features such as object boundaries may be smoothed out or inaccurately reconstructed, as illustrated in Fig.~\ref{fig:sharp edges}, ultimately reducing the geometric fidelity of the recovered depth map.

In contrast, our method leverages an explicit voxel grid representation, which models scene properties at discrete volumetric locations. This design offers enhanced flexibility in encoding abrupt changes in geometry and enables more accurate recovery of sharp surface transitions. Furthermore, the incorporation of surface color loss and distortion loss in our training framework enforces surface consistency while preserving local geometric details, achieving a favorable balance between smoothness and accuracy.

\begin{figure}
	\centering
	%\fbox{\rule{0pt}{2in} \rule{0.9\linewidth}{0pt}}
	\includegraphics[width=0.9\linewidth]{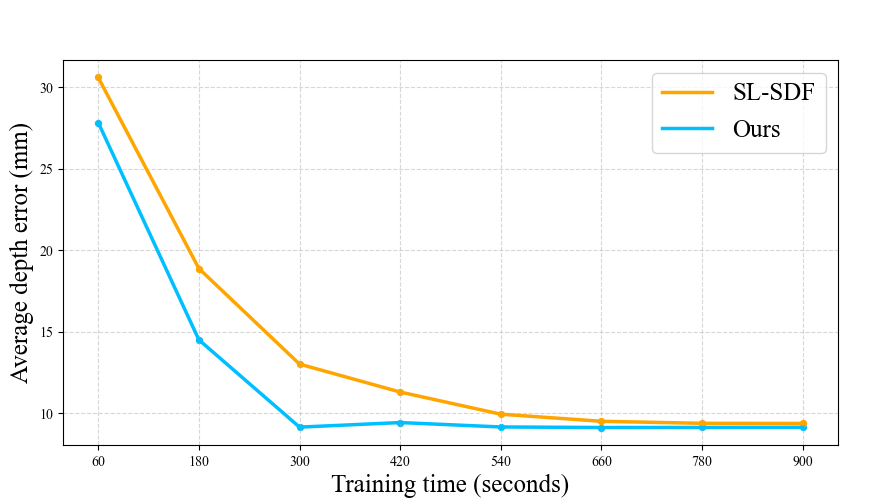}
	
	\caption{We adopt the same set of six projected patterns as used in SL-SDF for a fair comparison. Our method achieves convergence within 5 minutes of training, whereas SL-SDF requires more than 14 minutes. Despite the significantly reduced training time, our method attains comparable or superior accuracy.}
	\label{fig:comp with SL-SDF}
\end{figure}

In order to demonstrate the advantages of our voxel grid training schedule for structured-light-based depth recovery, we make further comparisons with SL-SDF~\cite{qiao2024depth}, which uses signed distance fields (SDFs) for a similar procedure. We explore the relationship between training time and depth estimation accuracy for both methods. As shown in Fig.~\ref{fig:comp with SL-SDF}, our method requires significantly less training time to achieve the same level of accuracy of recovered depths. This efficiency can be attributed to several factors, with one of the main reasons being the inherent differences in iteration speed between explicit and implicit scene representations. Specifically, with the same ray sampling batch size, each iteration in our method is approximately 20 times faster than in SL-SDF.

\begingroup
\setlength{\tabcolsep}{10pt} % Default value: 6pt
\renewcommand{\arraystretch}{1.5} % Default value: 1
\begin{table*}[htbp]
\centering
\caption{Ablation studies to analyze the impact of the rendered color loss, distortion loss, and surface color loss on the accuracy of depth recovery}
\label{tab:ab}
\begin{tabular}{l|cccc|cccc}
\toprule
\multicolumn{1}{c|}{\multirow{2}{*}{Losses}}                                                     & \multicolumn{4}{c|}{synthetic scenes}           & \multicolumn{4}{c}{real scenes}                 \\
\multicolumn{1}{c|}{}                                                                            & MAE(mm) & $O(0.1)$ & $O(0.5)$ & $O(1)$ & MAE(mm) & $O(0.1)$ & $O(0.5)$ & $O(1)$ \\ \hline
$\mathcal{L}_{\text {p}}$                                                                    & 35.844  & 0.21     & 0.12     & 0.11    & 38.712  & 7.13     & 4.69     & 1.91  \\
$\mathcal{L}_{\text {p}} + \mathcal{L}_{\text {d-t}}$                                 & 27.210  & 0.15     & 0.12     & 0.11   & 33.948  & 6.74     & 3.81     & 1.49    \\
$\mathcal{L}_{\text {p}} + \mathcal{L}_{\text {s}}$                                    & \colorbox[HTML]{FFCE93}{18.573}  & \colorbox[HTML]{FFCE93}{0.10}     & \colorbox[HTML]{FFCE93}{0.07}     & \colorbox[HTML]{FFCE93}{0.06}     & \colorbox[HTML]{FFCE93}{14.394}  & \colorbox[HTML]{FFCE93}{3.38}     & \colorbox[HTML]{FFCE93}{1.51}     & \colorbox[HTML]{FFCE93}{0.98}   \\
$\mathcal{L}_{\text {p}} + \mathcal{L}_{\text {d}} + \mathcal{L}_{\text {s}}$ & \colorbox[HTML]{FD6864}{\textbf{13.767}}  & \colorbox[HTML]{FD6864}{\textbf{0.08}}     & \colorbox[HTML]{FD6864}{\textbf{0.06}}     & \colorbox[HTML]{FD6864}{\textbf{0.06}}    & \colorbox[HTML]{FD6864}{\textbf{9.153}}   & \colorbox[HTML]{FD6864}{\textbf{2.87}}     & \colorbox[HTML]{FD6864}{\textbf{0.92}}     & \colorbox[HTML]{FD6864}{\textbf{0.47}}   \\ 
\bottomrule
\end{tabular}
\end{table*}
\endgroup

\subsection{Ablation Studies}

To evaluate the effectiveness of the individual loss components introduced in Section~\ref{sec:losses}, we conduct a series of ablation experiments using different combinations of loss functions on our synthetic benchmark scenes. Note that the photometric loss serves as the foundation of our optimization framework. Without it, the network fails to converge to a meaningful solution.

As shown in Table~\ref{tab:ab}, while the photometric loss alone enables the recovery of coarse geometric structures, it tends to overlook fine-scale surface variations, resulting in over-smoothed depth maps and missing details, especially near sharp edges and object boundaries. The inclusion of the surface color loss addresses this issue by encouraging the consistency between the predicted surface point color and the observed color, thus reinforcing the network’s ability to model subtle geometric details more faithfully.

Moreover, the distortion loss, which promotes a unimodal volume density distribution along each ray, helps to eliminate ambiguous or noisy depth estimates caused by multiple semi-transparent surfaces or low-frequency variations. By enforcing a single dominant depth response per ray, it significantly enhances the accuracy and stability of the final depth map. Together, the three loss terms work synergistically to improve both global structure recovery and local geometric fidelity.

\begin{figure*}
\centering
\includegraphics[width=0.9\linewidth]{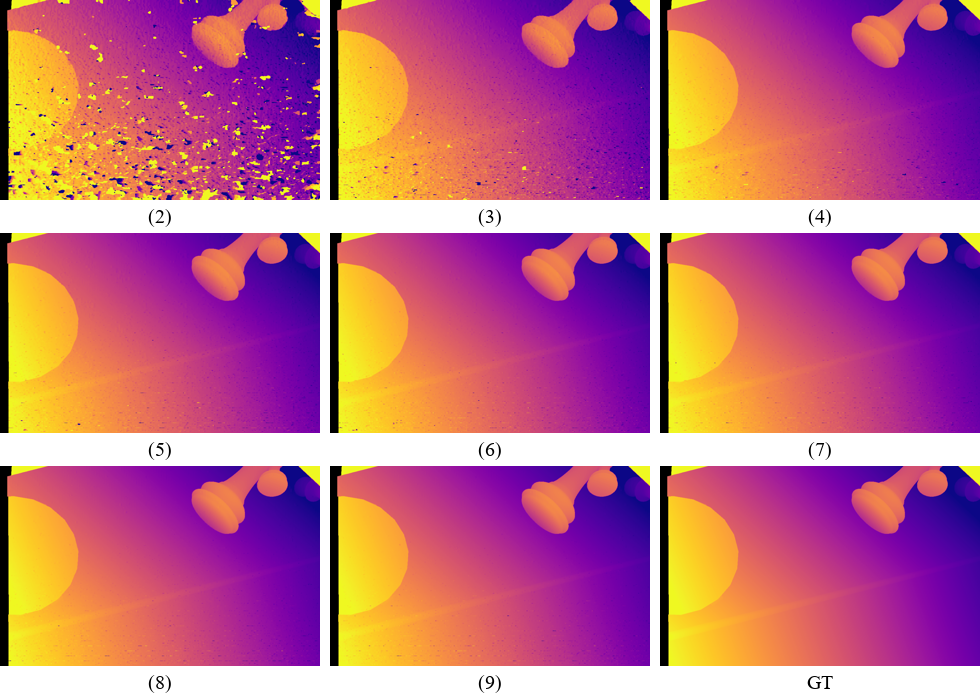}
\caption{Qualitative results under different numbers of projected patterns. The number below each image indicates how many patterns were used during both scene capture and voxel grid training. The rightmost image shows the ground truth depth map for reference.}
\label{fig:patnum}
\end{figure*}%

We also examine the effect of the number of projected patterns on the performance of our method. In the full configuration, we generate a set of nine 2D binary patterns using unit square lengths of 20, 10, and 5 pixels. To evaluate the influence of pattern count, we progressively reduce the number of projected patterns by following a structured removal order: one pattern of length 20, followed by one of length 10, then one of length 5, and repeating this cycle (i.e., removal sequence: 20, 10, 5, 20, 10, 5, 20). This strategy ensures a balanced degradation of spatial information across different scales, allowing us to assess how various frequency components contribute to the depth estimation process. We verify 10 synthetic scenes for this experiment and summarize the results in Table~\ref{tab:patnum} and Fig.~\ref{fig:patnum}. The results demonstrate that our voxel-based training process achieves satisfactory performance with as few as 6 patterns, with no significant improvement observed by further increasing the number of patterns. Notably, even with only four projected patterns, our approach still delivers promising results, highlighting its robustness and efficiency under limited input conditions.

\begingroup
\begin{table}
\centering
\caption{Ablation Study on the Number of Projected Patterns in Our Framework}
\label{tab:patnum}
\begin{tabular}{c|ccccccc}
\toprule
Pattern Count & 2 & 3 & 4 & 5 & 6 & 7 & 8\\
\midrule
MAE(mm) & 203.9 & 31.7 & 18.3 & 16.9 & 15.3 & 15.2 & 15.0 \\
\bottomrule
\end{tabular}
\end{table}
\section{Conclusion}
\endgroup

In this paper, we propose a novel framework for depth recovery in structured light systems using 3D voxel grids. Our approach centered on training a density voxel grid to represent the geometry of the captured scene, leveraging constraints from projected patterns to guide the training schedule through a fully differentiable rendering process. Upon convergence, we used volume density queried from the trained voxel grid to obtain a depth map through a similar rendering approach. A key advantage of our approach is that it completely eliminates the need for traditional correspondence search in the image space, thereby avoiding dependence on potentially error-prone matching algorithms. Experimental results demonstrate that our method achieves competitive, and in many cases superior, performance compared to conventional matching-based approaches and deep-learning-based techniques, while requiring the same or even fewer projected patterns. When compared with similar rendering-based methods, which use implicit functions to represent geometries, our approach demonstrates superior depth estimation accuracy while having faster training speed.

\bibliographystyle{IEEEtran}
\bibliography{arxiv}

% Generated by IEEEtran.bst, version: 1.14 (2015/08/26)
\begin{thebibliography}{10}
\providecommand{\url}[1]{#1}
\csname url@samestyle\endcsname
\providecommand{\newblock}{\relax}
\providecommand{\bibinfo}[2]{#2}
\providecommand{\BIBentrySTDinterwordspacing}{\spaceskip=0pt\relax}
\providecommand{\BIBentryALTinterwordstretchfactor}{4}
\providecommand{\BIBentryALTinterwordspacing}{\spaceskip=\fontdimen2\font plus
\BIBentryALTinterwordstretchfactor\fontdimen3\font minus \fontdimen4\font\relax}
\providecommand{\BIBforeignlanguage}[2]{{%
\expandafter\ifx\csname l@#1\endcsname\relax
\typeout{** WARNING: IEEEtran.bst: No hyphenation pattern has been}%
\typeout{** loaded for the language `#1'. Using the pattern for}%
\typeout{** the default language instead.}%
\else
\language=\csname l@#1\endcsname
\fi
#2}}
\providecommand{\BIBdecl}{\relax}
\BIBdecl

\bibitem{keselman2017intel}
L.~Keselman, J.~Iselin~Woodfill, A.~Grunnet-Jepsen, and A.~Bhowmik, ``Intel realsense stereoscopic depth cameras,'' in \emph{Proc. IEEE Conf. Comput. Vis. Pattern Recognit. workshops}, 2017, pp. 1--10.

\bibitem{batlle1998recent}
J.~Batlle, E.~Mouaddib, and J.~Salvi, ``Recent progress in coded structured light as a technique to solve the correspondence problem: a survey,'' \emph{Pattern Recognit.}, vol.~31, no.~7, pp. 963--982, 1998.

\bibitem{cai2020single}
Z.~Cai, G.~Pedrini, W.~Osten, X.~Liu, and X.~Peng, ``Single-shot structured-light-field three-dimensional imaging,'' \emph{Opt. Lett.}, vol.~45, no.~12, pp. 3256--3259, 2020.

\bibitem{salvi2010state}
J.~Salvi, S.~Fernandez, T.~Pribanic, and X.~Llado, ``A state of the art in structured light patterns for surface profilometry,'' \emph{Pattern Recognit.}, vol.~43, no.~8, pp. 2666--2680, 2010.

\bibitem{young2007coded}
M.~Young, E.~Beeson, J.~Davis, S.~Rusinkiewicz, and R.~Ramamoorthi, ``Coded structured light,'' in \emph{Proc. IEEE Conf. Comput. Vis. Pattern Recognit.}, 2007, pp. 1--8.

\bibitem{mirdehghan2018optimal}
P.~Mirdehghan, W.~Chen, and K.~N. Kutulakos, ``Optimal structured light a la carte,'' in \emph{Proc. IEEE Conf. Comput. Vis. Pattern Recognit.}, 2018, pp. 6248--6257.

\bibitem{jia2024adaptivestereo}
T.~Jia, X.~Li, X.~Yang, S.~Lin, Y.~Liu, and D.~Chen, ``Adaptivestereo: Depth estimation from adaptive structured light,'' \emph{Opt. Laser Technol.}, vol. 169, p. 110076, 2024.

\bibitem{kawasaki2008dynamic}
H.~Kawasaki, R.~Furukawa, R.~Sagawa, and Y.~Yagi, ``Dynamic scene shape reconstruction using a single structured light pattern,'' in \emph{Proc. IEEE Conf. Comput. Vis. Pattern Recognit.}, 2008, pp. 1--8.

\bibitem{sagawa2009dense}
R.~Sagawa, Y.~Ota, Y.~Yagi, R.~Furukawa, N.~Asada, and H.~Kawasaki, ``Dense 3d reconstruction method using a single pattern for fast moving object,'' in \emph{IEEE 12th Proc. IEEE Int. Conf. Comput. Vis.}, 2009, pp. 1779--1786.

\bibitem{sagawa2011dense}
R.~Sagawa, H.~Kawasaki, S.~Kiyota, and R.~Furukawa, ``Dense one-shot 3d reconstruction by detecting continuous regions with parallel line projection,'' in \emph{Proc. IEEE Int. Conf. Comput. Vis.}, 2011, pp. 1911--1918.

\bibitem{fu2023fast}
J.~Fu, Y.~Zhang, Y.~Li, J.~Li, and Z.~Xiong, ``Fast 3d reconstruction via event-based structured light with spatio-temporal coding,'' \emph{Opt. Exp.}, vol.~31, no.~26, pp. 44\,588--44\,602, 2023.

\bibitem{johari2021depthinspace}
M.~M. Johari, C.~Carta, and F.~Fleuret, ``Depthinspace: Exploitation and fusion of multiple video frames for structured-light depth estimation,'' in \emph{Proc. IEEE Int. Conf. Comput. Vis.}, 2021, pp. 6039--6048.

\bibitem{riegler2019connecting}
G.~Riegler, Y.~Liao, S.~Donne, V.~Koltun, and A.~Geiger, ``Connecting the dots: Learning representations for active monocular depth estimation,'' in \emph{Proc. IEEE Conf. Comput. Vis. Pattern Recognit.}, 2019, pp. 7624--7633.

\bibitem{qiao2022tide}
R.~Qiao, H.~Kawasaki, and H.~Zha, ``Tide: Temporally incremental disparity estimation via pattern flow in structured light system,'' \emph{IEEE Robotics and Automation Letters}, vol.~7, no.~2, pp. 5111--5118, 2022.

\bibitem{schreiberhuber2022gigadepth}
S.~Schreiberhuber, J.-B. Weibel, T.~Patten, and M.~Vincze, ``Gigadepth: Learning depth from structured light with branching neural networks,'' in \emph{Proc. Comput. Vis.–ECCV: 17th Eur. Conf.}\hskip 1em plus 0.5em minus 0.4em\relax Springer, 2022, pp. 214--229.

\bibitem{garbin2021fastnerf}
S.~J. Garbin, M.~Kowalski, M.~Johnson, J.~Shotton, and J.~Valentin, ``Fastnerf: High-fidelity neural rendering at 200fps,'' in \emph{Proc. IEEE Int. Conf. Comput. Vis.}, 2021, pp. 14\,346--14\,355.

\bibitem{yu2021plenoctrees}
A.~Yu, R.~Li, M.~Tancik, H.~Li, R.~Ng, and A.~Kanazawa, ``Plenoctrees for real-time rendering of neural radiance fields,'' in \emph{Proc. IEEE Int. Conf. Comput. Vis.}, 2021, pp. 5752--5761.

\bibitem{chen2022fast}
S.~Chen, B.~Yan, X.~Sang, D.~Chen, P.~Wang, Z.~Yang, X.~Guo, and C.~Zhong, ``Fast virtual view synthesis for an 8k 3d light-field display based on cutoff-nerf and 3d voxel rendering,'' \emph{Opt. Exp.}, vol.~30, no.~24, pp. 44\,201--44\,217, 2022.

\bibitem{mildenhall2021nerf}
B.~Mildenhall, P.~P. Srinivasan, M.~Tancik, J.~T. Barron, R.~Ramamoorthi, and R.~Ng, ``Nerf: Representing scenes as neural radiance fields for view synthesis,'' \emph{Communications of the ACM}, vol.~65, no.~1, pp. 99--106, 2021.

\bibitem{yu2021pixelnerf}
A.~Yu, V.~Ye, M.~Tancik, and A.~Kanazawa, ``pixelnerf: Neural radiance fields from one or few images,'' in \emph{Proc. IEEE Conf. Comput. Vis. Pattern Recognit.}, 2021, pp. 4578--4587.

\bibitem{jin2024reliable}
Z.~Jin, Z.~Xu, H.~Feng, Q.~Li, and Y.~Chen, ``Reliable image dehazing by nerf,'' \emph{Opt. Exp.}, vol.~32, no.~3, pp. 3528--3550, 2024.

\bibitem{niemeyer2022regnerf}
M.~Niemeyer, J.~T. Barron, B.~Mildenhall, M.~S. Sajjadi, A.~Geiger, and N.~Radwan, ``Regnerf: Regularizing neural radiance fields for view synthesis from sparse inputs,'' in \emph{Proc. IEEE Conf. Comput. Vis. Pattern Recognit.}, 2022, pp. 5480--5490.

\bibitem{wang2021neus}
P.~Wang, L.~Liu, Y.~Liu, C.~Theobalt, T.~Komura, and W.~Wang, ``Neus: Learning neural implicit surfaces by volume rendering for multi-view reconstruction,'' in \emph{Proc. Int. Conf. Neural Inf. Process. Syst.}, 2021.

\bibitem{yan2021continual}
Z.~Yan, Y.~Tian, X.~Shi, P.~Guo, P.~Wang, and H.~Zha, ``Continual neural mapping: Learning an implicit scene representation from sequential observations,'' in \emph{Proc. IEEE Int. Conf. Comput. Vis.}, 2021, pp. 15\,782--15\,792.

\bibitem{yariv2021volume}
L.~Yariv, J.~Gu, Y.~Kasten, and Y.~Lipman, ``Volume rendering of neural implicit surfaces,'' \emph{Proc. Int. Conf. Neural Inf. Process. Syst.}, vol.~34, pp. 4805--4815, 2021.

\bibitem{shandilya2023neural}
A.~Shandilya, B.~Attal, C.~Richardt, J.~Tompkin, and M.~O'toole, ``Neural fields for structured lighting,'' in \emph{Proc. IEEE Int. Conf. Comput. Vis.}, 2023, pp. 3512--3522.

\bibitem{qiao2024depth}
R.~Qiao, H.~Kawasaki, and H.~Zha, ``Depth reconstruction with neural signed distance fields in structured light systems,'' in \emph{3DV}, 2024, pp. 770--779.

\bibitem{mirdehghan2024turbosl}
P.~Mirdehghan, M.~Wu, W.~Chen, D.~B. Lindell, and K.~N. Kutulakos, ``Turbosl: dense accurate and fast 3d by neural inverse structured light,'' in \emph{Proc. IEEE Conf. Comput. Vis. Pattern Recognit.}, 2024, pp. 25\,067--25\,076.

\bibitem{lorensen1998marching}
W.~E. Lorensen and H.~E. Cline, ``Marching cubes: A high resolution 3d surface construction algorithm,'' in \emph{Seminal graphics: pioneering efforts that shaped the field}, 1998, pp. 347--353.

\bibitem{scharstein2003high}
D.~Scharstein and R.~Szeliski, ``High-accuracy stereo depth maps using structured light,'' in \emph{Proc. IEEE Conf. Comput. Vis. Pattern Recognit.}, vol.~1.\hskip 1em plus 0.5em minus 0.4em\relax IEEE, 2003, pp. I--I.

\bibitem{posdamer1982surface}
J.~L. Posdamer and M.~D. Altschuler, ``Surface measurement by space-encoded projected beam systems,'' \emph{Computer graphics and image processing}, vol.~18, no.~1, pp. 1--17, 1982.

\bibitem{aliaga2008photogeometric}
D.~G. Aliaga and Y.~Xu, ``Photogeometric structured light: A self-calibrating and multi-viewpoint framework for accurate 3d modeling,'' in \emph{Proc. IEEE Conf. Comput. Vis. Pattern Recognit.}, 2008, pp. 1--8.

\bibitem{weinmann2011multi}
M.~Weinmann, C.~Schwartz, R.~Ruiters, and R.~Klein, ``A multi-camera, multi-projector super-resolution framework for structured light,'' in \emph{Proc. Int. Conf. 3D Imag., Modeling, Process., Visualizat. Transmiss.}, 2011, pp. 397--404.

\bibitem{lei2013design}
Y.~Lei, K.~R. Bengtson, L.~Li, and J.~P. Allebach, ``Design and decoding of an m-array pattern for low-cost structured light 3d reconstruction systems,'' in \emph{Proc. IEEE Int. Conf. Image Process.}\hskip 1em plus 0.5em minus 0.4em\relax IEEE, 2013, pp. 2168--2172.

\bibitem{taguchi2012motion}
Y.~Taguchi, A.~Agrawal, and O.~Tuzel, ``Motion-aware structured light using spatio-temporal decodable patterns,'' in \emph{Proc. Comput. Vis.–ECCV: 12th Eur. Conf.}\hskip 1em plus 0.5em minus 0.4em\relax Springer, 2012, pp. 832--845.

\bibitem{wang20133d}
Y.~Wang, J.~I. Laughner, I.~R. Efimov, and S.~Zhang, ``3d absolute shape measurement of live rabbit hearts with a superfast two-frequency phase-shifting technique,'' \emph{Opt. Exp.}, vol.~21, no.~5, pp. 5822--5832, 2013.

\bibitem{zuo2013high}
C.~Zuo, Q.~Chen, G.~Gu, S.~Feng, F.~Feng, R.~Li, and G.~Shen, ``High-speed three-dimensional shape measurement for dynamic scenes using bi-frequency tripolar pulse-width-modulation fringe projection,'' \emph{Opt. Lasers Eng.}, vol.~51, no.~8, pp. 953--960, 2013.

\bibitem{sundar2022single}
V.~Sundar, S.~Ma, A.~C. Sankaranarayanan, and M.~Gupta, ``Single-photon structured light,'' in \emph{Proc. IEEE Conf. Comput. Vis. Pattern Recognit.}, 2022, pp. 17\,865--17\,875.

\bibitem{koninckx2006real}
T.~P. Koninckx and L.~Van~Gool, ``Real-time range acquisition by adaptive structured light,'' \emph{IEEE Trans. Pattern Anal. Mach. Intell.}, vol.~28, no.~3, pp. 432--445, 2006.

\bibitem{fanello2016hyperdepth}
S.~R. Fanello, C.~Rhemann, V.~Tankovich, A.~Kowdle, S.~O. Escolano, D.~Kim, and S.~Izadi, ``Hyperdepth: Learning depth from structured light without matching,'' in \emph{Proc. IEEE Conf. Comput. Vis. Pattern Recognit.}, 2016, pp. 5441--5450.

\bibitem{fanello2017ultrastereo}
S.~R. Fanello, J.~Valentin, C.~Rhemann, A.~Kowdle, V.~Tankovich, P.~Davidson, and S.~Izadi, ``Ultrastereo: Efficient learning-based matching for active stereo systems,'' in \emph{Proc. IEEE Conf. Comput. Vis. Pattern Recognit.}\hskip 1em plus 0.5em minus 0.4em\relax IEEE, 2017, pp. 6535--6544.

\bibitem{zhang2018activestereonet}
Y.~Zhang, S.~Khamis, C.~Rhemann, J.~Valentin, A.~Kowdle, V.~Tankovich, M.~Schoenberg, S.~Izadi, T.~Funkhouser, and S.~Fanello, ``Activestereonet: End-to-end self-supervised learning for active stereo systems,'' in \emph{Proc. Comput. Vis.–ECCV: 15th Eur. Conf.}, 2018, pp. 784--801.

\bibitem{maturana2015voxnet}
D.~Maturana and S.~Scherer, ``Voxnet: A 3d convolutional neural network for real-time object recognition,'' in \emph{IEEE/RSJ International Conference on Intelligent Robots and Systems}.\hskip 1em plus 0.5em minus 0.4em\relax IEEE, 2015, pp. 922--928.

\bibitem{schwarz2022voxgraf}
K.~Schwarz, A.~Sauer, M.~Niemeyer, Y.~Liao, and A.~Geiger, ``Voxgraf: Fast 3d-aware image synthesis with sparse voxel grids,'' \emph{Proc. Int. Conf. Neural Inf. Process. Syst.}, vol.~35, pp. 33\,999--34\,011, 2022.

\bibitem{sun2022direct}
C.~Sun, M.~Sun, and H.-T. Chen, ``Direct voxel grid optimization: Super-fast convergence for radiance fields reconstruction,'' in \emph{Proc. IEEE Conf. Comput. Vis. Pattern Recognit.}, 2022, pp. 5459--5469.

\bibitem{mildenhall2020nerf}
B.~Mildenhall, P.~P. Srinivasan, M.~Tancik, J.~T. Barron, R.~Ramamoorthi, and R.~Ng, ``Nerf: Representing scenes as neural radiance fields for view synthesis,'' \emph{Proc. Comput. Vis.–ECCV: 16th Eur. Conf.}, pp. 405--421, 2020.

\bibitem{barron2021mip}
J.~T. Barron, B.~Mildenhall, M.~Tancik, P.~Hedman, R.~Martin-Brualla, and P.~P. Srinivasan, ``Mip-nerf: A multiscale representation for anti-aliasing neural radiance fields,'' in \emph{Proc. IEEE Int. Conf. Comput. Vis.}, 2021, pp. 5855--5864.

\bibitem{barron2022mip}
J.~T. Barron, B.~Mildenhall, D.~Verbin, P.~P. Srinivasan, and P.~Hedman, ``Mip-nerf 360: Unbounded anti-aliased neural radiance fields,'' in \emph{Proc. IEEE Conf. Comput. Vis. Pattern Recognit.}, 2022, pp. 5470--5479.

\bibitem{garg2016unsupervised}
R.~Garg, V.~K. Bg, G.~Carneiro, and I.~Reid, ``Unsupervised cnn for single view depth estimation: Geometry to the rescue,'' in \emph{Proc. Comput. Vis.–ECCV: 14th Eur. Conf.}\hskip 1em plus 0.5em minus 0.4em\relax Springer, 2016, pp. 740--756.

\bibitem{darmon2022improving}
F.~Darmon, B.~Bascle, J.-C. Devaux, P.~Monasse, and M.~Aubry, ``Improving neural implicit surfaces geometry with patch warping,'' in \emph{Proc. IEEE Conf. Comput. Vis. Pattern Recognit.}, 2022, pp. 6260--6269.

\bibitem{xiangli2022bungeenerf}
Y.~Xiangli, L.~Xu, X.~Pan, N.~Zhao, A.~Rao, C.~Theobalt, B.~Dai, and D.~Lin, ``Bungeenerf: Progressive neural radiance field for extreme multi-scale scene rendering,'' in \emph{Proc. Comput. Vis.–ECCV: 17th Eur. Conf.}, 2022, pp. 106--122.

\bibitem{chang2015shapenet}
A.~X. Chang, T.~Funkhouser, L.~Guibas, P.~Hanrahan, Q.~Huang, Z.~Li, S.~Savarese, M.~Savva, S.~Song, H.~Su \emph{et~al.}, ``Shapenet: An information-rich 3d model repository,'' \emph{arXiv preprint arXiv:1512.03012}, 2015.

\bibitem{wu2020high}
Z.~Wu, W.~Guo, Y.~Li, Y.~Liu, and Q.~Zhang, ``High-speed and high-efficiency three-dimensional shape measurement based on gray-coded light,'' \emph{Photonics Res.}, vol.~8, no.~6, pp. 819--829, 2020.

\end{thebibliography}

% \vfill

\end{document}